\documentclass[accepted]{melba}

\usepackage{amsmath,amsfonts}
\usepackage{multirow}
\usepackage{color, colortbl}
\definecolor{Gray}{gray}{0.9}
\usepackage{xspace}
\usepackage{url}
\usepackage{xcolor}

\usepackage{booktabs}

\usepackage{amsmath}
\usepackage{algorithm, algorithmic}
\usepackage[]{changes}

\newcommand{\stepone}{task-specific supervised training\xspace}
\newcommand{\steptwo}{task-expansive prompt-based learning\xspace}
\newcommand{\stepthree}{task-agnostic self-supervised learning\xspace}

\newcommand{\ie}{i.e.\@\xspace}

\newcommand{\usteptwo}{Task-general Prompt-based Learning\xspace}
\newcommand{\ustepthree}{Task-agnostic Self-supervised Learning\xspace}

\melbaid{2025:006}  
\doi{doi.org/10.59275/j.melba.2025-86a6}
\melbaauthors{Gu, Dong, Yang and Mazurowski}  
\email{hanxue.gu@duke.edu}
\volume{2}
\firstpageno{88}  
\melbayear{2025}  
\datesubmitted{2024-08-22}  
\datepublished{2025-04-29}  

\ShortHeadings{Medical Image Segmentation with Foundation Models}{Gu, Dong, Yang and Mazurowski}

\title{How to build the best medical image segmentation algorithm using foundation models: a comprehensive empirical study with Segment Anything Model}

\author{
	\firstname Hanxue \surname Gu\aff{1}$^{*}$\orcid{0000-0003-2622-753X},
	\firstname Haoyu \surname Dong\aff{1}$^{*}$\orcid{0000-0002-5132-0341},
        \firstname Jichen \surname Yang\aff{1}\orcid{0000-0001-8781-5461},
        \firstname Maciej A. \surname Mazurowski\aff{1,2,3,4}\orcid{0000-0003-4202-8602}
}
\affiliations{
	\num 1 \addr Department of Electrical and Computer Engineering, Duke University, US\\
	\num 2 \addr Department of Biostatistics \& Bioinformatics, Duke University, US\\
	\num 3 \addr Department of Computer Science, Duke University, US\\
        \num 4 \addr Department of Radiology, Duke University, US\\
        $^{*}$These authors contributed equally to this work.
}

\abstract{
Automated segmentation is a fundamental medical image analysis task, which enjoys significant advances due to the advent of deep learning. While foundation models have been useful in natural language processing and some vision tasks for some time, the foundation model developed with image segmentation in mind - Segment Anything Model (SAM) - has been developed only recently and has shown similar promise. However, there are still no systematic analyses or ``best-practice'' guidelines for optimal fine-tuning of SAM for medical image segmentation. 
This work summarizes existing fine-tuning strategies with various backbone architectures, model components, and fine-tuning algorithms across 18 combinations and evaluates them on 17 datasets covering all common radiology modalities. 
Our study reveals that (1) fine-tuning SAM leads to slightly better performance than previous segmentation methods, (2) fine-tuning strategies that use parameter-efficient learning in both the encoder and decoder are superior to other strategies, (3) network architecture has a small impact on final performance, and (4) further training SAM with self-supervised learning can improve final model performance. 
We also demonstrate the ineffectiveness of some methods popular in the literature and further expand our experiments into few-shot and prompt-based settings.
Lastly, we released our code and MRI-specific fine-tuned weights, which consistently obtained superior performance over the original SAM, at ~\url{https://github.com/mazurowski-lab/finetune-SAM}. }


\begin{document}

\twocolumn[\maketitle]

\section{Introduction}
\label{sec:introduction}
Medical image segmentation refers to identifying the region of interest, typically anatomical structures or tumors, in an image.
This process is essential in the field of medical image analysis, contributing significantly to the effectiveness of computer-aided diagnosis ~\citep{al2018fully}.
The introduction of deep learning techniques has led to a variety of automated segmentation methods ~\citep{liu2020extracting, weng2023deep}, further enhancing the precision and efficiency of analyzing medical images. 
Despite the progress made, it is still difficult to determine the most effective method for developing an optimal segmentation model, especially with the recent introduction of vision foundation models dedicated to segmentation. 
Therefore, in our study, we aim to explore and define the optimal strategies for developing automated segmentation algorithms specifically designed for medical images and with a focus on foundation models. 


In the era of deep learning, a common practice for automated medical image segmentation is to directly train UNet~\citep{ronneberger2015u} or its variants~\citep{siddique2021u,wang2022medical} on a specific dataset and task.
Recently, there has been growing interest in customizing vision foundation models for specific segmentation tasks ~\citep{zhang2023customized, gu2024segmentanybone}.
Foundation models are typically transformer-based~\citep{vaswani2017attention} neural networks pre-trained on extremely large datasets, allowing them to generalize to a wide range of tasks and datasets.
The Segment Anything Model (SAM) ~\citep{kirillov2023segment} was proposed as the first large model designed specifically for segmentation tasks.
SAM employs a prompt-based training method, aiming to generate segmentation masks for target objects in response to user-provided prompts, e.g., points or bounding boxes. It has demonstrated comparable or even superior zero-shot performance to fully supervised models in processing natural images. 
However, SAM performance is less impressive when applied directly to medical images ~\citep{mazurowski2023segment, Deng2023SegmentAM, huang2024segment}, and the requirement of prompts makes its direct usage with medical imaging segmentation tasks challenging. 

Removing the prompt requirement in SAM is relatively straightforward, as we can use ``dummy'' prompt embeddings as inputs during fine-tuning, a strategy verified by concurrent work ~\citep{zhang2023customized}.
However, the approaches proposed for adapting SAM to medical imaging tasks vary widely. Some works focused on adapting SAM's decoder only ~\citep{yue2023surgicalsam}, while others chose to modify the entire network ~\citep{zhang2023customized}. Some works introduced an additional pre-training stage ~\citep{MedSAM} using multiple medical images, while others usually do not. Though some prior survey work ~\citep{lee2024foundation, zhang2023towards} have attempted to list different strategies and categorize them into various groups, questions remain regarding (1) which fine-tuned strategy is the most effective, (2) whether these models can achieve better performance than traditional UNets, and (3) whether using additional data is helpful.

\begin{figure*}[t]
    \centering
    \includegraphics[width=0.95\textwidth]{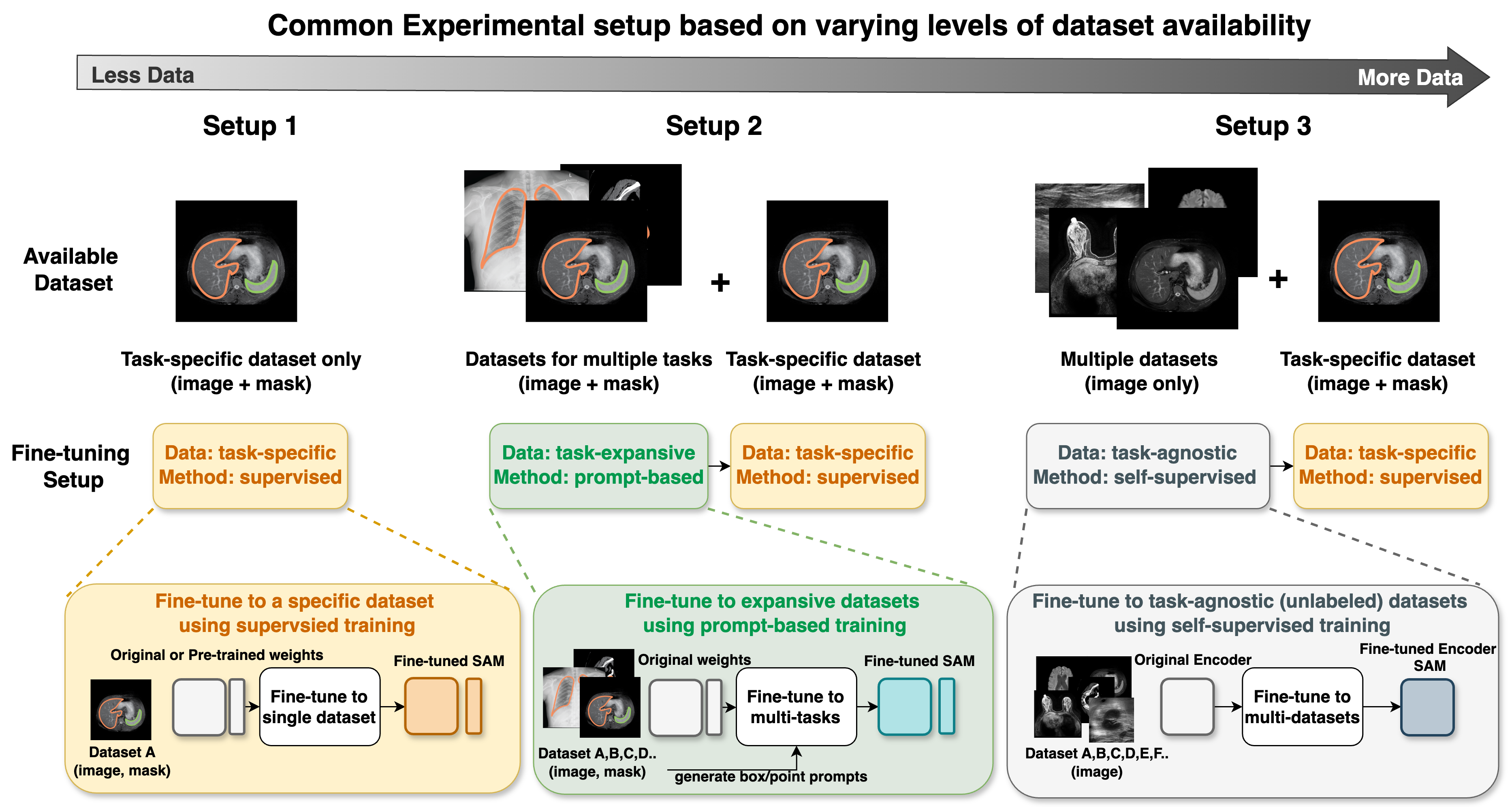}
    \caption{Overview of general fine-tuning strategies based on different levels of dataset availability.}
    \label{fig:overview}
\end{figure*}

In this work, we aim to answer these questions by considering three common dataset availability scenarios in the medical imaging field: (1) only a single labeled dataset; (2) multiple labeled datasets for different tasks; and (3) multiple labeled and unlabeled datasets. 
The first scenario reflects the most common deep learning scenario of adapting SAM to a dataset of interest, and the latter two scenarios offer the opportunity to first incorporate broad medical domain knowledge into SAM using supervised or self-supervised training before adapting it to a specific dataset.
Figure \ref{fig:overview} summarizes the strategies and experiment setups we include.
To avoid terminology confusion, we name these strategies as \textbf{\stepone, \steptwo}, and \textbf{\stepthree} respectively.

When considering the design choice during \stepone, our work aligns with several concurrent works ~\citep{lee2024foundation,zhang2023towards,zhang2024segment}.
However, our work differs from theirs in that we 
conduct comprehensive experiments over $17$ medical imaging datasets across various modalities for a fair comparison, rather than simply compiling disparate results, and provide an organized and new catalog that could summarize existing approaches.
The choice of each category is shown in Figure \ref{fig:strategy} and will be discussed in Section \ref{sec:method}.
Our category definition reflects the common design choices when building a segmentation algorithm from SAM, and we can categorize most existing works based on the proposed criteria, shown in Table \ref{tab:summary}.

When incorporating extra medical knowledge into SAM, we explore two main setups of pre-training approaches: \steptwo and \stepthree, depending on the availability of labels.
Note that our objective is not to propose a novel pre-training strategy, as such strategies are often constrained by the availability of computation resources and datasets. Instead, our focus is on determining if additional pre-training with more available medical data is helpful and, if so, what the best combination of pre-training and task-specific fine-tuning is.
Therefore, we adopt the strategies used during SAM's development, and one of the most commonly applied pre-training strategies in this field.
Namely, we use the same interactive training pipeline where prompts are simulated from the masks for \steptwo; and masked autoencoding (MAE)~\citep{he2022masked} as the objective for \stepthree. 
We evaluate these approaches on the same 17 datasets. 

Our study is extended to two crucial scenarios when deploying a segmentation algorithm for annotation purposes: few-shot learning and interactive segmentation.
The first scenario reflects the case of annotating a new dataset with only a limited number of annotations, while the latter examines the effectiveness of using tools to assist the annotation process. 
In both scenarios, we limit the evaluation datasets to MRI. When conducting few-shot learning, we limit the number of training samples to 5 and re-examine the effectiveness of setups 1-3 described in Figure \ref{fig:overview}. 
In the interactive segmentation scenario, we focus on \stepone with either boxes or points as input prompts.

In conclusion, we systematically study different approaches to fine-tune SAM in the medical imaging field.
Our paper investigates an important question: What is the best practice to develop an automated segmentation algorithm based on foundation models? We find the answers depend on the availability of the datasets and consider three scenarios: (1) only a single labeled dataset; (2) multiple labeled datasets (with different objects of interest); and (3) multiple labeled and unlabeled datasets. We propose \stepone, \steptwo followed by \stepone, and \stepthree followed by \stepone as the solution to each scenario, respectively. 
By conducting systematic experiments on each strategy, we observe several useful guidelines and summarize them in Section \ref{sec:conclusion}.

\section{Method}
In this section, we first recapitulate SAM's structure. Next, we describe the selected methods under different data availability scenarios.

\subsection{SAM and removal of prompts}
\begin{figure*}[t]
    \centering
    \includegraphics[width=0.95\textwidth]{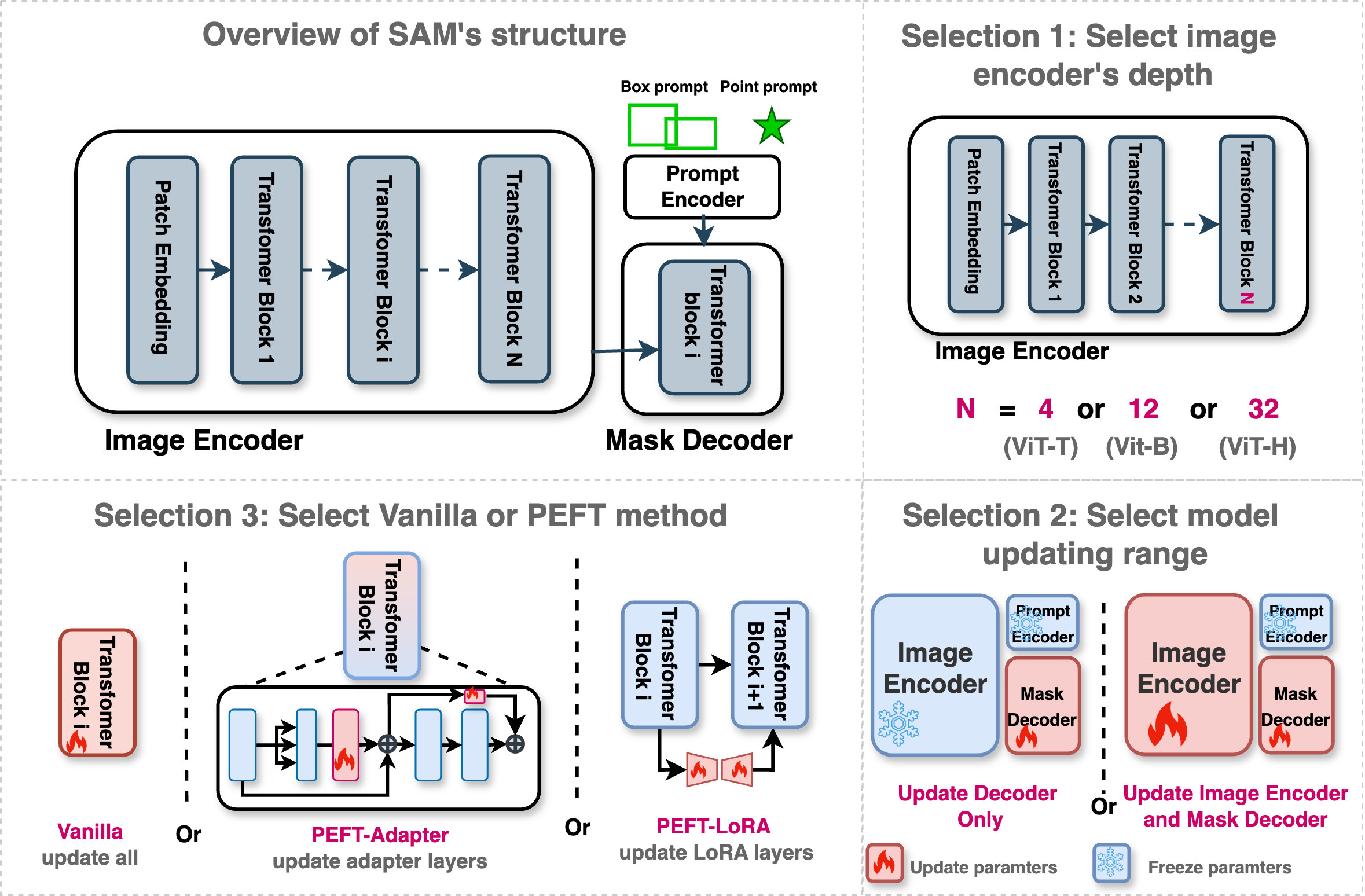}
    \caption{Visualization of {the overview of the SAM's architecture and } task-specific fine-tuning architectures selected in our study: including 3 encoder architecture $\times$ 2 model updating range $\times$ 3 vanilla/PEFT methods = 18 choices.}
    \label{fig:strategy}
\end{figure*}

To begin with, we first briefly present an overview of the Segment Anything Model (SAM) ~\citep{kirillov2023segment} before introducing the fine-tuning strategies.
SAM consists of three components: (1) image encoder, (2) prompt encoder, and (3) mask decoder, as illustrated in Figure \ref{fig:strategy} of the overview of SAM's architecture. The image encoder, built on a robust Vision Transformer (ViT) framework, effectively converts a 2D image into a latent feature representation. Notably, in SAM's original architecture, the image encoder offers flexibility in model size, with options including ViT-H(uge), ViT-L(arge), and ViT-B(ase) ~\citep{dosoViTskiy2020image}. The prompt encoder is flexible, accommodating different prompt types such as point, box, or mask prompts. Depending on the input type, it generates either sparse embeddings for point/box prompts or dense embeddings for mask prompts. Lastly, the mask decoder is a transformer decoder using two-way cross-attention that integrates image and prompt embeddings to produce a multi-channel mask output, each channel representing varying confidence levels.
SAM is trained on SA-1B, a dataset collected by the same team behind its development, which contains more than 1B masks from 11M natural images. During training, first, with equal probabilities, box and point prompts are generated. Point prompts are sampled uniformly from the ground truth mask, and box prompts are taken as the ground truth mask's bounding box, with random noise added in each coordinate. The objective is the combination of focal loss ~\citep{ross2017focal} and dice loss ~\citep{milletari2016v}.

\subsubsection{Interactive segmentation to automated segmentation}
\label{sec:auto}
When converting SAM to auto-mode (without prompts), we can simply use the default embeddings generated by using ``None'' as inputs to the prompt encoder, \ie
\begin{verbatim}
    sparse_embeddings, dense_embeddings = 
    sam.prompt_encoder(points=None, 
    boxes=None, masks=None),
\end{verbatim}
where the ``sparse\_embeddings'' and ``dense\_embeddings'' serve as inputs to SAM's mask decoder.
Empirical evidence shows the effectiveness of this simple approach ~\citep{zhang2023customized}. 
More advanced methods, including adding a detection network to automatically generate box prompts or a self-prompt module into SAM's structure ~\citep{wu2023self,gao2023desam,shaharabany2023autosam}, are feasible but will introduce extra complexity into SAM's structure, hampering our analysis on the effectiveness of fine-tuning strategies.
Moreover, we believe our work offers a potentially effective strategy, \ie, feeding the generated prompts to the fine-tuned SAM instead of the original one, for this research direction. 

\subsection{Variables in \stepone}
\label{sec:method}
We consider three aspects in \stepone that can affect the performance, as shown in Figure \ref{fig:strategy}. The choice of each category is discussed next.

\subsubsection{Image Encoder Architecture}
In this work, we select ViT-B and ViT-H from SAM's provided checkpoints because they represent the most efficient and effective versions of SAM's encoder, respectively.
Additionally, we explore a more compact option: ViT-T(iny). This backbone is introduced in MobileSAM ~\citep{zhang2023faster}, where the authors distill the knowledge from SAM's pre-trained ViT-H to ViT-T, a customized vision transformer with only $1\%$ of the network parameters, and achieve an inference speed that is $50$ times faster compared to the original SAM. The lightweight nature can widen the application of SAM, ease the computational requirements, and, more importantly, raise interesting research questions, such as whether distillation can maintain or even lead to the superior performance of SAM during fine-tuning.


\subsubsection{Model Component}
When selecting which set(s) of SAM's parameters to update, the most straightforward way is to select one or multiple components from SAM. 
As discussed in Section \ref{sec:auto}, we use the default prompt embeddings and thus can keep the prompt encoder as is. 

Some methods state the effectiveness of maintaining the encoder's weights and only adapting the lightweight mask decoder of SAM's architecture, as shown in Table \ref{tab:summary}.
These approaches are grounded in the understanding that SAM's encoder, trained on a comprehensive collection of natural images, is sufficiently powerful and knowledgeable in the general imaging domain, with zero-shot ability to perform feature extraction. Updating the lightweight decoder can enhance performance in specific applications.
Also, it is suggested that fine-tuning the encoder of SAM would demand considerably higher computational resources.
However, due to the significant appeal difference between natural images and medical images, it remains unclear if SAM's encoder could effectively capture the distinctive features of medical images without any modifications to the image encoder.
Therefore, we consider these two choices: \textit{mask decoder only} and \textit{mask decoder plus image encoder}.

\begin{figure*}
\centering
\includegraphics[width=0.98\textwidth]{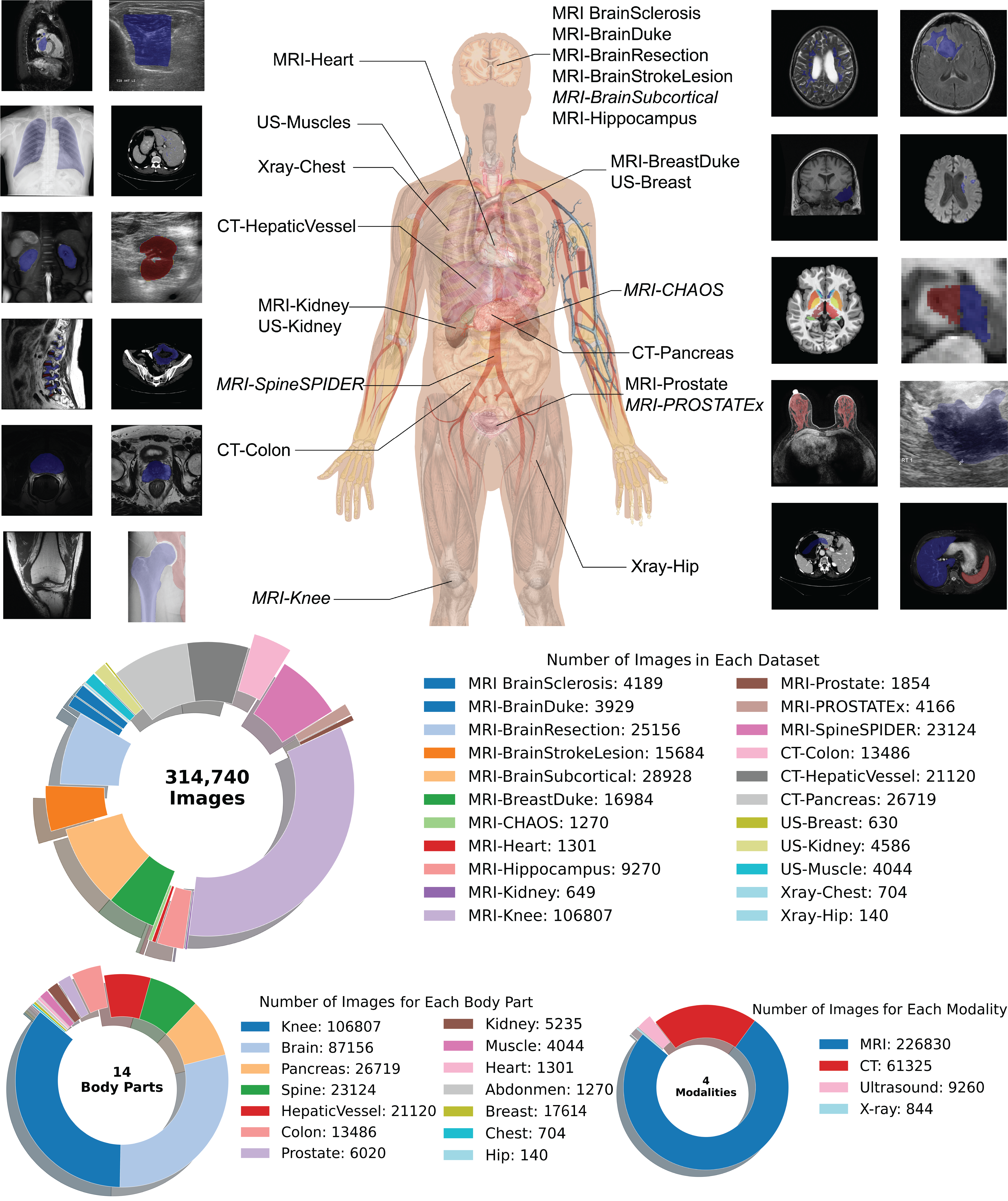}
\caption{A visual summary of the collected datasets, including the visual example from each dataset and the number of images under different criteria.}
\label{fig:data}
\end{figure*}

\begin{table*}[h]
\centering
\footnotesize
\begin{tabular}{lllcc}  
\hline
   Category & Methods & Dataset & Backbone \\
   \hline
   \multirow{10}{*}{Decoder + Vanilla} 
   & SAM for medical images?~\citep{huang2024segment} & COSMOS 1050K & ViT-H / ViT-B\\
   & Generalist Vision Foundation~\citep{shi2023generalist} & Multiple & ViT-H \\
   & How to efficiently~\citep{hu2023efficiently} & ACDC~\citep{bernard2018deep} & ViT-H \\
   & All-in-SA~\citep{cui2023all} & Monuseg~\citep{kumar2019multi} & ViT-H \\
   & AdaptiveSAM~\citep{paranjape2023adaptivesam} & Multiple & ViT-B \\
   & SkinSAM~\citep{hu2023skinsam} & HAM10000~\citep{tschandl2018ham10000} & ViT-B \\
   & DeSAM~\citep{gao2023desam} & Multi-Prostate & ViT-H & \\
   & SurgicalSAM~\citep{yue2023surgicalsam} & EndoVis~\citep{allan20202018} & ViT-H \\
   & SAM3D~\citep{bui2023sam3d} & Multiple & ViT-B \\
   & SAM Fewshot~\citep{xie2024sam} & Multiple & N/A \\
   \hline
   \multirow{1}{*}{Decoder + LoRA} 
   & BLO-SAM~\citep{zhang2024blo} & Multiple & ViT-B \\
   \hline
   \multirow{2}{*}{Encoder + Vanilla}
   & AutoSAM~\citep{shaharabany2023autosam} & Multiple & ViT-H \\
   & SAMUS~\citep{lin2023samus} & Multiple & N/A \\
   \hline
   \multirow{2}{*}{En/Decoder + Vanilla}
   & Polyp-SAM~\citep{li2023polyp} & Multi-Colonoscopy & ViT-L / ViT-B \\
   & CellViT~\citep{horst2023cellViT} & Multi-nuclei & ViT-H \\
   \hline
   \multirow{3}{*}{En/Decoder + LoRA} 
   & Customized SAM~\citep{zhang2023customized} & Multi-Organ & ViT-H / ViT-B \\
   & MA-SAM~\citep{chen2023ma} & BTVC~\citep{fang2020multi} & ViT-H \\
   & Cheap Lunch~\citep{feng2023cheap} & BraTS 18~\citep{menze2014multimodal} & ViT-B \\
   \hline 
   \multirow{9}{*}{En/Decoder + Adapter} 
   & Medical SAM Adapter~\citep{wu2023medical} & BTVC~\citep{fang2020multi} & ViT-H \\
   & SAM Fails to Segment?~\citep{chen2023sam} & kvasir-SEG~\citep{jha2020kvasir} & ViT-H \\
   & Learnable Ophthalmology SAM~\citep{qiu2023learnable} & Multiple & N/A \\
   & Auto-Prompting SAM~\citep{li2023auto} & Multiple & LightWeight ViT \\
   & SAM-Med2D~\citep{cheng2023sam} & SAM-Med2D~\citep{cheng2023sam} & ViT-B \\
   & 3DSAM-adapter~\citep{gong20233dsam} & Multi-CT & ViT-B \\
   & SAM-Path~\citep{zhang2023sam} & Multiple & LightWeight ViT \\
   & Self-Sampling Meta~\citep{leng2024self} & ABD-30~\citep{landman2015miccai} & ViT-B \\
   & Cross-modality~\citep{shi2023cross} & MRI & ViT-B \\
   \hline
\end{tabular} 
\caption{Summary of existing fine-tuning methods, grouped by the component-method combinations. En/Decoder means both the encoder and decoder are optimized.}
\label{tab:summary}
\end{table*}


\subsubsection{Vanilla or Parameter-efficient Fine-tuning Method}
When conducting model fine-tuning or transfer learning, the easiest and most intuitive approach is to update all model parameters involved when feeding medical domain data, which is known as vanilla fine-tuning ~\citep{azizi2022robust}.

Recently, parameter-efficient fine-tuning (PEFT) techniques have gained increasing attention. Compared with vanilla fine-tuning, PEFT keeps most of the network parameters fixed and only updates a small percentage, \ie less than $5\%$, of them.
Such a strategy significantly reduces computational cost and is beneficial in preventing both catastrophic forgetting and overfitting ~\citep{dutt2023parameter}.
The simplest PEFT techniques involve selecting and fine-tuning a subset of the parameters from the original pre-trained model.
These selective methods concentrate on identifying the optimal parameters for fine-tuning, aiming to strike an ideal balance between parameter efficiency and overall model performance.
For instance, BitFit ~\citep{zaken2021bitfit} proposed fine-tuning only the bias term in a transformer-based network, and Tinytl ~\citep{cai2020tinytl} shares a similar idea but focuses on CNN-based networks. 
Touvron et al. ~\citep{touvron2022three} have demonstrated the effectiveness of selectively fine-tuning only the attention layers in Vision Transformers (ViTs), whereas alternative approaches suggest that exclusively optimizing the normalization layers ~\citep{basu2023strong}.

Rather than updating certain selected layers, recent research ~\citep{li2022cross} indicates that incorporating extra modules into the large models appears to be more beneficial because it reconfigures models towards a downstream task rather than solely updating network parameters.
Among these additional module methods, Adapters, which were originally introduced for learning multi-domain representations ~\citep{rebuffi2017learning,rebuffi2018efficient}, have emerged as a particularly prevalent choice. 
In addition to the conventional Adapter block ~\citep{houlsby2019parameter}, Low-Rank Adaptation (LoRA), known for its efficient low-rank matrices added into self-attention layers that require even fewer parameters, has also gained popularity ~\citep{hu2021lora}. 
Additionally, AdaptFormer ~\citep{chen2022adaptformer}, which integrates a scalable Adapter into the multi-layer perceptron (MLP) layer of the attention block specifically for vision transformers, has shown that adding just 2\% extra parameters can surpass the performance of fully fine-tuned networks across various natural imaging benchmarks.
In our study, we aim to investigate the effectiveness of applying PEFT for fine-tuning SAM within the medical imaging domain. Considering that Adapter and LoRA are the most popular PEFT methods adopted in LLMs and vision foundation transformers ~\citep{xu2023parameter}, we choose to include them, as well as the simplest Vanilla, in our experiments. 
{The implementation details of Adapter and LoRA are presented in Figure \ref{fig:strategy}, Selection 3. Specifically, Adapter introduces two additional layers in each transformer block, and LoRA adds trainable pairs of rank decomposition matrices in parallel to existing blocks. Following ~\citep{hu2021lora}, we set the rank of the additional matrices to 4, which provides a good balance between performance and computational efficiency.}
In summary, we select the following methods:
\begin{enumerate}
    \item Vanilla - a ``baseline'' approach that updates all parameters.
    \item Adapter / LoRA - these approaches involve adding extra layers with original parameters fixed. 
\end{enumerate}

\subsubsection{Summary}
To sum up, we choose the following combinations to fine-tune SAM's performance for specific datasets or tasks, as illustrated in Figure \ref{fig:strategy}: (1) Variations in the image encoder's architecture, including ViT-H, ViT-B, and ViT-T from MobileSAM ~\citep{zhang2023faster}. (2) Different approaches to the fine-tuning scope of the network, encompassing both the encoder and decoder updates, as well as the decoder only. (3) Diverse parameter updating settings, comprising Vanilla, PEFT with Adapter blocks, and PEFT with LoRA blocks.
By integrating these three variables, we present a total of $3 \times 2 \times 3= \textbf{18}$ distinct architectures for fine-tuning SAM on \stepone setup.

\begin{table*}[t]
\centering
{\scriptsize
\begin{tabular}{l|c|c|c}
    Alias in our paper    & Description and citation & \# of images & \# of masks\\
    \hline
    \rowcolor{Gray}
    \multicolumn{4}{l}{Datasets used for dataset-specific fine-tuning and pre-training:} \\ 
    \hline
    MRI-BrainSclerosis (MRI-Scler.) & Brain Sclerosis with Lesion Segmentation ~\citep{muslim2022brain} & $4189$ & $1813$ \\
    MRI-BrainDuke (MRI-Brain) & Brain with FLAIR Abnormality Segmentation Masks~\citep{buda2019brain} & $3929$ & $1372$ \\
    MRI-BrainResection (MRI-Resect.) & EPISURG Segmentation on the Resection CaViTy~\citep{perez2020episurg} & $25156$ & $5377$ \\
    MRI-BrainStrokeLesion (MRI-Stroke.) & ISLES 2022: Stroke Lesion Segmentation~\citep{hernandez2022isles} & $15684$ & $4827$ \\
    MRI-BreastDuke (MRI-Breast) & Breast Cancer Screening~\citep{saha2018machine,Lew2024APA} & $16984$ & $16984$ \\
    MRI-Heart & Medical Segmentation Decathlon Cardiac~\citep{antonelli2022medical} & $1301$ & $1301$ \\
    MRI-Hippocampus (MRI-Hippo.) & Medical Segmentation Decathlon Hippocampus~\citep{antonelli2022medical} & $9270$ & $6622$ \\
    MRI-Kidney & T2-weighted Kidney MRI~\citep{daniel2021automated} & $649$ & $647$ \\
    MRI-Prostate (MRI-Prost.) & Original Multi-Parametric MRI Images of Prostate~\citep{lemaitre2016original} & $1854$ & $1854$ \\
    CT-Colon & Medical Segmentation Decathlon Colon~\citep{antonelli2022medical} & $13486$ & $1285$ \\
    CT-HepaticVessel (CT-vessel) & Medical Segmentation Decathlon Hepatic Vessel~\citep{antonelli2022medical} & $21120$ & $13042$ \\
    CT-Pancreas (CT-Panc.) & Medical Segmentation Decathlon Pancreas~\citep{antonelli2022medical} & $26719$ & $8792$ \\
    US-Breast & Breast Ultrasound Images Dataset~\citep{al2019deep} & $630$ & $630$ \\
    US-Kidney & Kidney Segmentation in Ultrasound Images~\citep{song2022ct2us} & $4586$ & $4585$ \\
    US-Muscle & Transverse Musculoskeletal Ultrasound images~\citep{marzola2021deep} & $4044$ & $4043$ \\
    Xray-Chest & Montgomery County and Shenzhen Chest X-ray Dataset~\citep{jaeger2014two} & $704$ & $704$ \\
    Xray-Hip & X-ray Images of the Hip Joints~\citep{gut_x-ray_2021} & $140$ & $140$ \\
    \hline
    \rowcolor{Gray}
    \multicolumn{4}{l}{Datasets used for pre-training only:} \\ 
    \hline
    MRI-BrainSubcortical & 14 Brain Sub-cortical Structures~\citep{sivaswamy2021sub} & $28928$ & $5891$ \\
    MRI-CHAOS & Segmentation of Liver, Kidneys and Spleen from MRI data~\citep{kavur2021chaos} & $1270$ & $992$ \\
    MRI-Knee & MRNet Knee MRI Examinations~\citep{bien2018deep} & $106807$ & Not used \\
    MRI-ProstateX & ProstateX Dataset of Prostate MR Studies~\citep{armato2018prostatex} & $4166$ & $3195$ \\
    MRI-SpineSPIDER & SPIDER - Lumbar Spine Segmentation in MRI~\citep{van2023lumbar} & $23124$ & $12763$ \\
\end{tabular}
}
\caption{Description of the collected datasets. The first $17$ datasets are used throughout the paper for training and evaluation, and the remaining $5$ are used exclusively for supervised or self-supervised pre-training.}
\label{tab:data}
\end{table*}

\subsection{\usteptwo}
\label{section:supervised}
When multiple \textbf{labeled} datasets are available, an approach to adapting SAM to the medical imaging field involves additional pre-training, \ie starting with SAM's pre-trained weights and continuing the learning process with a compilation of the medical datasets, followed by \stepone, as is similar in ~\citep{cheng2023sam, MedSAM}.
The procedure is illustrated in Figure \ref{fig:overview} setup $2$. 
The motivation for adopting a two-step training approach, rather than utilizing automated fine-tuning on multiple datasets, lies in the fact that various datasets, despite featuring medical images of the same anatomical area and being captured using the same imaging technique, may concentrate on entirely distinct objectives.
For example, two brain MRI datasets might each focus on segmenting different entities, such as the hippocampus in one and tumors in the other. Applying automated segmentation to these datasets could introduce ambiguity and make the training challenging.

Therefore, we use a prompt-based segmentation setup during \steptwo, which involves providing prompts to identify the targeted area of interest interactively.
This approach addresses the issue of the above-mentioned ambiguity.
Furthermore, this choice conceptually aligns with the second phase of SAM's second-stage training process ~\citep{kirillov2023segment}, allowing it to be considered as an extra step of supervised training specifically tailored for medical images within the SAM framework.

To simulate a prompt, we first randomly selected a type of object if the ground truth mask contains multiple objects.
Then, for each contiguous region of the selected object, we generate a box or a point prompt with equal probability.
The generation of box prompts follows SAM's strategy: a tight box containing the target box is created first, followed by some random noises added to both locations.
The random noises are generated with the standard deviation equal to $10\%$ of the box coordinates and to a maximum of $20$ pixels.
For point generation, we follow ~\citep{mazurowski2023segment} to first compute the distance of each pixel to its closest edge within the object region and sample a pixel with a distance above the median distance.
This operation allows us to sample a point prompt around the center of the object.
Such a design choice differs from that of the SAM, which samples the first point prompt uniformly and subsequent points from the error region between the previous mask prediction and the ground-truth mask.
Since our ultimate goal is to learn an automated segmentation algorithm, we don't need an iterative process to refine predictions based on prompts.

Considering the differences across various medical imaging modalities, alongside the observation that images within the same modality often share similar visual features ~\citep{konz2024intrinsicproperties}, 
we emphasize the value of modality-specific dataset knowledge in pre-training, a practice also common in the literature ~\citep{tang2022self}. 
Specifically, we focus on supervised pre-training on MRI, which comprises most of the collected datasets.

\subsection{\ustepthree}
\label{section:self-supervised}
When using multiple \textbf{unlabeled} medical image datasets, we can utilize the same idea as in the last section to conduct additional pre-training before applying \stepone.
As the label information is unavailable in this scenario, we adapt the self-supervised learning (SSL) techniques, referred to as \stepthree. 
SSL refers to training a network without any human-collected labels. 
This is an effective strategy since the number of available datasets can increase dramatically.
Popular SSL methods generally fall into two categories: contrastive learning ~\citep{He2020MomentumCF, chen2020simple, park2020contrastive} or masked image modeling ~\citep{he2022masked, xie2022simmim}.
In the medical image field, further methods are tailored to fit the characteristics of medical images ~\citep{tang2022self}, including dimension discrepancy ~\citep{xie2022unimiss}, structure similarity ~\citep{jiang2023anatomical}, etc. 

Among these methods, we select the Masked Autoencoder (MAE) ~\citep{he2022masked} as our self-supervised pre-training strategy for three reasons: 
(1) The computation of CL is infeasible given the input size of SAM, \ie $1024\times 1024$;
(2) SAM also utilized an MAE pre-trained vision transformer as the initial model;
(3) There is increasing evidence showing the effectiveness of using MAE during medical image segmentation ~\citep{zhou2023self, zhuang2023advancing, wang2024fremim}.
MAE first divides the input images into non-overlapping patches and removes most of them. Its goal is to reconstruct the original inputs given the remaining ones.
For the same reason as in \steptwo, we also conduct training with MRIs only and evaluate all datasets.

\section{Dataset{s}}
We select 22 datasets from publicly available sources online, including TCIA (The Cancer Imaging Archive), Kaggle, Zendo, Grand Challenge, and Mendeley Data. Figure \ref{fig:data} presents a visual summary of the datasets used in this paper. In total, we collect $14$ magnetic resonance images (MRI) datasets, $3$ computed tomography (CT) datasets, $3$ ultrasound (US) datasets, and $2$ X-ray datasets. Our datasets cover images from $14$ main human anatomical locations. Examples of all $22$ datasets are also included in the same figure, in addition to the image count of each dataset, each anatomical location, and each modality. The details of each dataset are listed in Table \ref{tab:data}, including the citation, description, number of images, and number of masks. All datasets use expert-annotated segmentation masks, except for the MRI-Knee dataset, which is used exclusively for self-supervised pre-training. Therefore, the number of masks for this dataset is indicated as ``not used''.

Since SAM is a 2D-based segmentation network, 3D volumes are first extracted to 2D slices. 
We convert NIfTI, MHA, and DICOM volumes by normalizing all pixel values to 0-255 and further upsampling each slice to $ 1024 \times 1024$ using nearest-neighbor interpolation to meet SAM's required input size. The pixel values are repeated three times across the three channels for all grayscale images. 
We split each dataset into training, validation, and test sets by dividing patients into $7:1:2$ splits, accordingly.

In our study, we utilize these curated medical imaging datasets across various levels of data availability as follows:
\begin{enumerate}
    \item \textit{Specific Dataset Fine-tuning}: This involves datasets applied for fine-tuning on a single labeled dataset, including all datasets listed in the first section of Table \ref{tab:data}.
    \item  \textit{Prompt-based Supervised Training with Multiple Labeled Datasets}: This category includes all MRI datasets that come with available masks, utilized for supervised pre-training.
    \item  \textit{Self-Supervised Learning with Both Labeled and Unlabeled Datasets}: This level pertains to the use of all MRI datasets mentioned in our table for self-supervised learning, incorporating both labeled and unlabeled datasets.
\end{enumerate}

\begin{figure*}[t]
    \centering
    \includegraphics[width=0.98\textwidth]{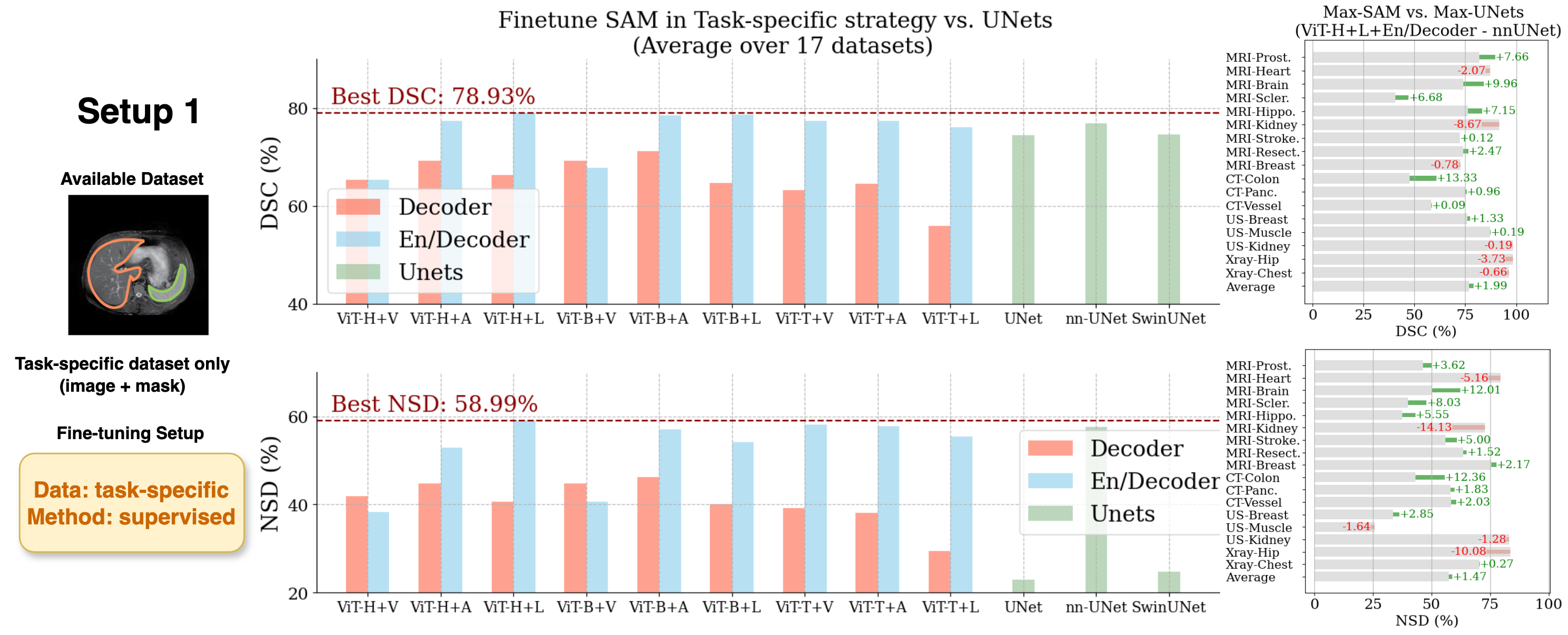}
    \caption{Left: Illustration of the dataset and training setting: \stepone;  Middle: Visualization of the performance of various \stepone strategies, including various Image-Encoder depth: 4 (ViT-T) or 12 (ViT-B) or 32 (ViT-H); updating all parameters (V) or use PEFT-Adapter (A) or PEFT-LoRA (L); fine-tuning Mask Decoder only (De) or Image Encoder + Mask Decoder (En/De), and UNet and its variances. Right: Difference between the best fine-tuned SAM model (ViT-H + En/Decoder + LoRA) and the best UNet-based model (nnUNet) across individual datasets.}
    \label{fig:task-specific}
\end{figure*}

\begin{table*}[t]
\centering
{\small
\begin{tabular}{ccc|c|c|c|c}
\multicolumn{3}{c|}{Model}                                                                                       & \multicolumn{1}{c|}{\# of param.} & \multicolumn{1}{c|}{\# of trainable param.} & \multicolumn{1}{c|}{GPU allocated (MB)} & Single iteration time (s) \\ \hline
\multicolumn{1}{c|}{\multirow{6}{*}{ViT-H}}     & \multicolumn{1}{c|}{\multirow{3}{*}{Decoder}}   & Vanilla & 640.95M                         & 3.92M (0.61\%)                                  & 22206                                   & 1.82                          \\
\multicolumn{1}{c|}{}                           & \multicolumn{1}{c|}{}                                & Adapter & 641.08M                         & 0.13M (0.021\%)                                                                   & 17210                                   & 1.88                          \\
\multicolumn{1}{c|}{}                           & \multicolumn{1}{c|}{}                                & LoRA    & 640.97M                         & 0.65M (0.10\%)                                                          & 17214                                   & 1.81                          \\ \cline{2-7} 
\multicolumn{1}{c|}{}                           & \multicolumn{1}{c|}{\multirow{3}{*}{En/Decoder}} & Vanilla & 640.95M                         & 640.95M (100.0\%)                                                                   & 99,208                                  & 5.2                           \\
\multicolumn{1}{c|}{}                           & \multicolumn{1}{c|}{}                                & Adapter & 650.93M                         & 9.98M (1.53\%)                                                               & 69,144                                  & 3.8                           \\
\multicolumn{1}{c|}{}                           & \multicolumn{1}{c|}{}                                & LoRA    & 641.63M                         & 1.31M  (0.20\%)                                                             & 68,540                                  & 3.82                          \\ \hline
\multicolumn{1}{c|}{\multirow{6}{*}{ViT-B}}     & \multicolumn{1}{c|}{\multirow{3}{*}{Decoder}}   & Vanilla & 93.60M                          & 3.92M (4.19\%)                                                               & 11158                                   & 0.44                          \\
\multicolumn{1}{c|}{}                           & \multicolumn{1}{c|}{}                                & Adapter & 93.73M                          & 0.13M (0.14\%)                                                             & 11532                                   & 0.45                          \\
\multicolumn{1}{c|}{}                           & \multicolumn{1}{c|}{}                                & LoRA    & 93.62M                          & 0.65M (0.69\%)                                                                  & 11530                                   & 0.44                          \\ \cline{2-7} 
\multicolumn{1}{c|}{}                           & \multicolumn{1}{c|}{\multirow{3}{*}{En/Decoder}} & Vanilla & 93.60M                          & 93.60M (100.0\%)                                                                           & 37306                                   & 1.16                          \\
\multicolumn{1}{c|}{}                           & \multicolumn{1}{c|}{}                                & Adapter & 97.28M                          & 3.68M (3.79\%)                                                                & 33454                                   & 1                             \\
\multicolumn{1}{c|}{}                           & \multicolumn{1}{c|}{}                                & LoRA    & 93.77M                          & 0.80M (0.86\%)                                                                & 30578                                   & 0.95                          \\ \hline
\multicolumn{1}{c|}{\multirow{6}{*}{ViT-T}} & \multicolumn{1}{c|}{\multirow{3}{*}{Decoder}}   & Vanilla & 9.99M                           & 3.92M (39.22\%)                                                                & 1404                                    & 0.08                          \\
\multicolumn{1}{c|}{}                           & \multicolumn{1}{c|}{}                                & Adapter & 10.12M                          & 0.13M (1.13\%)                                                              & 1398                                    & 0.09                          \\
\multicolumn{1}{c|}{}                           & \multicolumn{1}{c|}{}                                & LoRA    & 10.01M                          & 0.65M (6.49\%)                                                                  & 1400                                    & 0.08                          \\ \cline{2-7} 
\multicolumn{1}{c|}{}                           & \multicolumn{1}{c|}{\multirow{3}{*}{En/Decoder}} & Vanilla & 9.99M                           & 9.99M (100.0\%)                                                                            & 10102                                   & 0.25                          \\
\multicolumn{1}{c|}{}                           & \multicolumn{1}{c|}{}                                & Adapter & 10.31M                          & 0.32M (3.12\%)                                                                    & 4018                                    & 0.16                          \\
\multicolumn{1}{c|}{}                           & \multicolumn{1}{c|}{}                                & LoRA    & 10.04M                          & 0.79M (7.82\%)                                                           & 4036                                    & 0.15                         
\end{tabular}
}
    \caption{Specifics of parameters and computer resources occupied for 18 combinations of fine-tuning choices. GPU allocations are measured when the batch size is set to 4, and the time for forward and backward passes is measured on a single A6000.}
    \label{tab:param_table}
\end{table*}

\begin{figure*}[t]
    \centering
    \includegraphics[width=\textwidth]{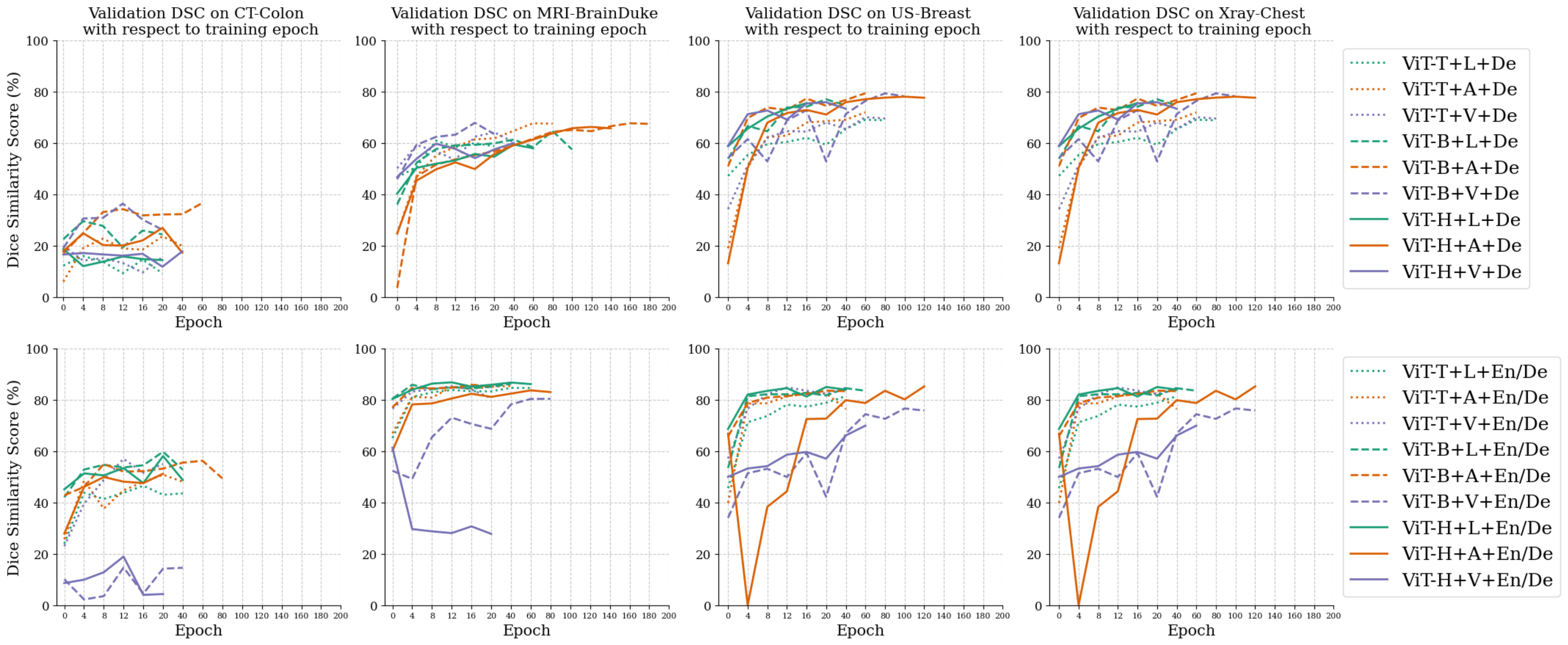}
    \caption{Validation {Dice Coefficient Score (DSC)} with respect to training epoch for different combinations of fine-tuning strategies. We select one representative dataset from each modality for visualization, and the behavior is consistent across different datasets. Note that the total number of epochs is not consistent because the training stops when the validation DSC is not improved for 20 epochs.} 
    \label{fig:vis_loss}
\end{figure*}

\section{Experiment{s}}
\label{sec:experiment}
In this section, we conduct various experiments to assess the effectiveness of fine-tuning strategies across multiple datasets and configurations. First and foremost, we analyze the behaviors of these strategies when conducting \stepone. Next, we examine the effectiveness of applying \steptwo or \stepthree before \stepone. These experiments are further extended to the few-shot learning and interactive segmentation setting. Throughout the paper, we use the Dice similarity score (DSC\%) and normalized surface distance (NSD\%) as the evaluation metric.

\subsection{Which fine-tuning combination is the best if applying \stepone only?}
\label{section:task-specfic}
\subsubsection{Implementation details}
When conducting \stepone, in addition to finding the most effective fine-tuning strategy, we also want to compare the performance against conventional UNet-based algorithms. 
Therefore, we have chosen three well-known and frequently used automated segmentation models for comparison: UNet ~\citep{ronneberger2015u}, nnUNet ~\citep{isensee2021nnu}, and SwinUNet ~\citep{cao2022swin}:
\begin{enumerate}
    \item \textit{UNet} serves as the baseline of the convolutional neural network-based segmentation algorithm.
    \item \textit{nnUNet} configures a UNet-based algorithm according to the specific attributes of the input datasets and approximates the performance of UNet with the optimal design choice.
    \item \textit{SwinUNet} replaces the encoder of UNet with a swin-transformer ~\citep{liu2021swin}.
\end{enumerate}

For PEFT with Adapter blocks, we implemented it based on ~\citep{wu2023self}, which added adapter blocks to SAM's attention blocks. In the image encoder, adapter blocks were added into transformer blocks right after the multi-attention head or within the MLP residual block with a scalable parameter ~\citep{chen2022adaptformer}. In the default setting, we added Adapter blocks at the first two and last two transformer blocks (layers) in the image encoder.
In the mask decoder, Adapters were added after the multi-head attention. 
For PEFT with LoRA blocks, we implemented it based on SAMed~\citep{zhang2023customized}, which inserted LoRA layers into the ``query'' and ``value'' projection layers of each transformer block in the image encoder and mask decoder. 

In the pre-processing step for our datasets, we removed all data lacking target objects, specifically images where the corresponding masks only consisted of the background, for both training and testing purposes. 
We also followed the same normalization mean and standard deviation as the original SAM.
We utilized dice loss and cross-entropy loss with equal weights.
For all experiments across various datasets, we adopted a basic learning rate of 1e-4 with a warmup of $200$ iterations and applied AdamW with $0.1$ weight decay for the remaining iterations. 
The batch size was set to $4$. 
Histogram equalization and color jitter were used as augmentations during training. 
The maximum epochs were set to $200$ epochs, incorporating an early stopping mechanism that halted training if there was no improvement in the highest validation DSC for $20$ consecutive epochs.
For settings that require heavy computing resources, such as ViT-H with the vanilla setting, we used Distributed Data Parallel (DDP) combined with model parallelism to split batches or the model across 2-3 GPUs. 
The specifics regarding the total number of parameters in the model, the number of parameters subject to updates for each combination, and the computing resources needed are detailed in Table \ref{tab:param_table}.

\subsubsection{Results}

Figure \ref{fig:task-specific} shows the performance of various task-specific fine-tuning approaches across each dataset and the performance of various UNet-based algorithms.
The detailed results of each architecture-component-method fine-tuning strategy over 17 datasets can be found in the Appendix. Note that we provide the numerical results for all figure-based results in the Appendix. 
By comparing the red and blue bars, we first observe an almost definite improvement (except in the ViT-B + Vanilla setting) in average performance when updating both the encoder and decoder parameters.
This discovery indicates that SAM's original image encoder, initially trained on a vast collection of natural images, is not well-suited for efficiently extracting features from medical images, aligned with previous studies ~\citep{mazurowski2023segment}. Therefore, freezing it and directly adapting the mask decoder for downstream tasks in the medical domain yields suboptimal performance.

When comparing the performances of Vanilla, Adapter, and LoRA, we observe little difference between the two PEFT methods (Adapter and LoRA). The rank between Vanilla and the PEFT methods depends on the choices of backbones. When using the standard-size backbone (ViT-H or ViT-B), we observe an absolute advantage of using PEFT techniques. For example, when comparing the middle three blue bars in Figure \ref{fig:task-specific}, we observe that using PEFT with Adapter or LoRA achieves $77.35\%$ and $78.93\%$ DSC, respectively, while vanilla fine-tuning yields only $65.39\%$ DSC. 
When comparing Adapter and LoRA, we observe that LoRA performs better when applied to a model with a larger size, \textit{i.e.,} ViT-H. This can be caused by the superior parameter efficiency of LoRA ~\citep{hu2021lora}.
We also conducted an additional study to insert the adapter block after each layer and observed little difference. 
However, the difference between vanilla fine-tuning and PEFT is significantly reduced when using ViT-T, the small backbone introduced by MobileSAM ~\citep{zhang2023faster}.
We hypothesize that the large gap between Vanilla and PEFT methods with a standard-size backbone is caused by the model collapse and knowledge-forgotten issues when fine-tuning all parameters of a large model ~\citep{dutt2023parameter, fu2023effectiveness}.
To support our hypothesis, we track the change in validation DSC on all datasets. Examples of each modality are shown in Figure \ref{fig:vis_loss}. 
It reveals that ViT-H + En/Decoder + Vanilla would cause a model collapse at an early stage. For instance, when applied to the MRI-BrainDuke dataset, this configuration results in a significant performance decrease, with a drop of approximately $35\%$ in DSC.

Next, our results show that the choice of encoder network architecture has a minimal impact on performance when all other decisions remain the same. For instance, under the same settings of updating both encoder and decoder with Adapter (EN/Decoder + Adapter), we observe similar DSC scores across different encoders: ViT-H, ViT-B, and ViT-T. 
This similarity in performance occurs despite ViT-H having a significantly larger encoder with $6.8$ times network parameters compared with ViT-B and $64$ times parameters compared with ViT-T, as listed in Table \ref{tab:param_table}. We deduce that while ViT-H excels in foundational settings for multiple segmentation tasks, it may be too parameter-rich for singular medical imaging segmentation tasks, given the typically smaller dataset sizes in the medical imaging field. 
Such property can also make the training difficult. As shown in Figure \ref{fig:vis_loss} top right, it takes $140$ epochs to train the ViT-H + En/Decoder + Adapter setting, which is approximately $30$ times longer when conducting the same setting with ViT-B, yet the latter gives slightly better performance ($80.53\%$ DSC vs. $81.21\%$ DSC when using ViT-H and ViT-B as the backbone respectively).
Given the uniform results across our 17 datasets, we deduce that using PEFT (Adapter or LoRA) to fine-tune SAM within a smaller framework offers more benefits. This strategy strikes an optimal balance between efficiency and performance, especially for medical image segmentation tasks where datasets typically range from a couple of hundred to tens of thousands of images.

Lastly, when comparing the results of automated segmentation by directly training on UNets, we observe that fine-tuning SAM with the En/Decoder generally yields better average performance. 
To further study the optimal performance across individual datasets, we select the best SAM-based method (ViT-H+LoRA+En/Decoder) vs. the best UNet-based method (nnUNet) in Figure \ref{fig:task-specific}, right block.
It appears that SAM excels in tasks where the objects of interest are complicated, such as MRI-Brain Sclerosis or CT-Colon. Conversely, nnUNet demonstrates higher precision in tasks with simpler targets, including Xray-Hip or MRI-Kidney.

\begin{figure*}[t]
    \centering
    \includegraphics[width=\textwidth]{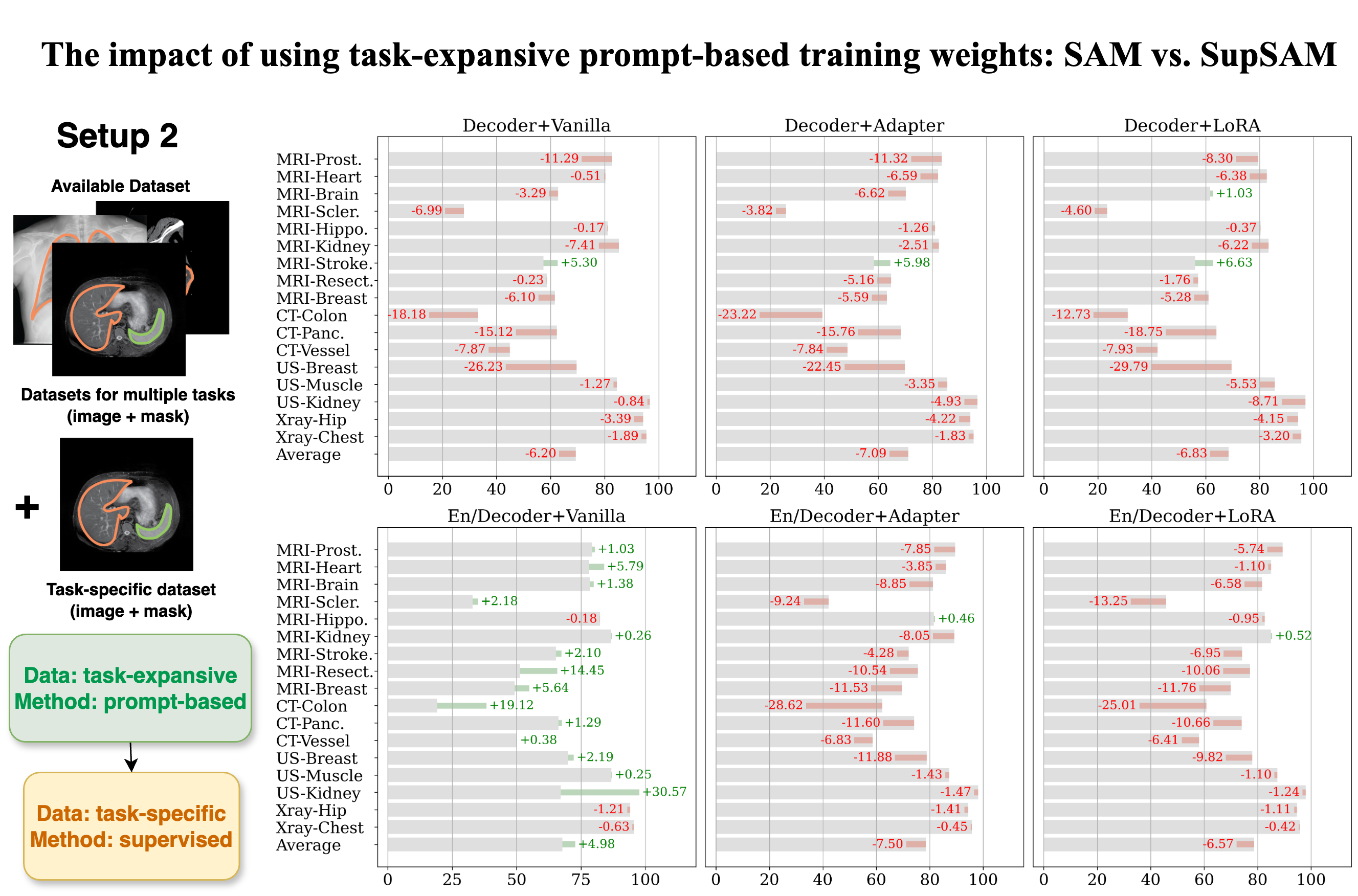}
    \caption{The {comparison} performance of applying various \stepone strategies to SAM {directly} (in gray), and {the approach of using \steptwo first when additional labeled datasets are available}. 
    The differences are represented as SupSAM - SAM, where green indicates a positive difference ($>$0) and red indicates a negative difference ($<$0), meaning that using SupSAM to initiate fine-tuning led to a performance increase or drop, respectively.}
    \label{fig:sup}
\end{figure*}

\begin{figure*}[t]
    \centering
    \includegraphics[width=\textwidth]{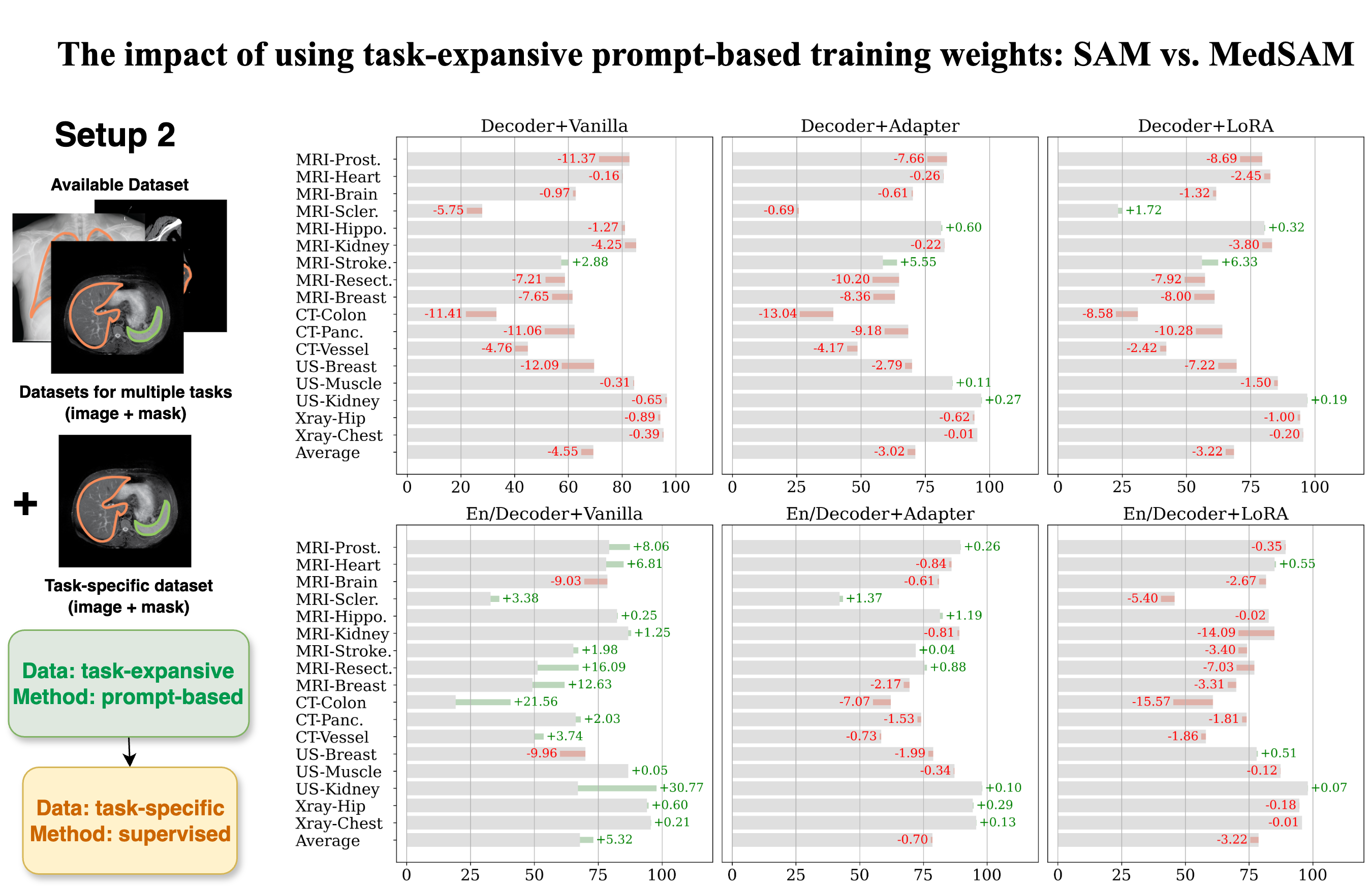}
    \caption{The {comparison} performance of applying various \stepone strategies to SAM {directly} (in gray), and the approach of using MedSAM ~\citep{MedSAM}, a variation of \steptwo first when additional labeled datasets are available. The differences are represented as MedSAM - SAM, where green indicates a positive difference ($>$0) and red indicates a negative difference ($<$0), meaning that using MedSAM to initiate fine-tuning led to a performance increase or drop, respectively.}
    \label{fig:medsam}
\end{figure*}

\subsection{Is \steptwo helpful for further automated segmentation?}
\label{sec:scp_experiment}
As discussed in Section \ref{section:supervised}, we utilize an interactive segmentation setting for \steptwo. In this section, we provide the implementation details and analysis of its performance.

\subsubsection{Implementation details}
When implementing \steptwo, we utilized a collection of datasets and their corresponding masks.
Our experimental setup focused on training with all MRI datasets listed in Table \ref{tab:data} except ``MRI-Knee''.
In this collection criteria, we include 13 MRI datasets, including 45k training images and non-empty mask pairs, and 7.9k evaluation images and mask pairs.
We selected ViT-B as the encoder architecture since (1) it achieved similar performance to ViT-H in the data-specific fine-tuning setting; (2) MobileSAM was a distilled version of SAM, which was not provided by the initial SAM and thus less representative. 

Under this multi-dataset supervised training setting, we configured a batch size of 3, utilizing DDP across 3 A6000 GPUs for training. 
The training was stopped when the validation performance did not improve for 5 epochs.
To train the network, we fixed the prompt encoder and updated the image encoder and mask decoder in a vanilla setting (same as SAM's supervised training stage).
We generated a box or point prompt for each image as described in Section \ref{section:supervised}.
Additionally, in this setting, we incorporated MedSAM ~\citep{MedSAM} as a parallel approach since its training algorithm was similar to the proposed \steptwo but with a different database.

In this experimental setup, our methodology follows a two-phase procedure, depicted in Figure \ref{fig:overview}. Setup 2: (1) Phase 1: \steptwo. (2) Phase 2: \stepone for each dataset, beginning from pre-trained weights.
We name the weights after phase 1 as \textbf{SupSAM}.
The implementation details in the task-specific setting were identical to those of the previous section (Section \ref{section:task-specfic}). 
In phase 2, we had 2 model component choices (En/Decoder or Decoder only) and 3 Vanilla/PEFT choices in the automated setting.  We evaluated the final automated segmentation performance after phase 2.

\begin{figure*}[t]
    \centering
    \includegraphics[width=\textwidth]{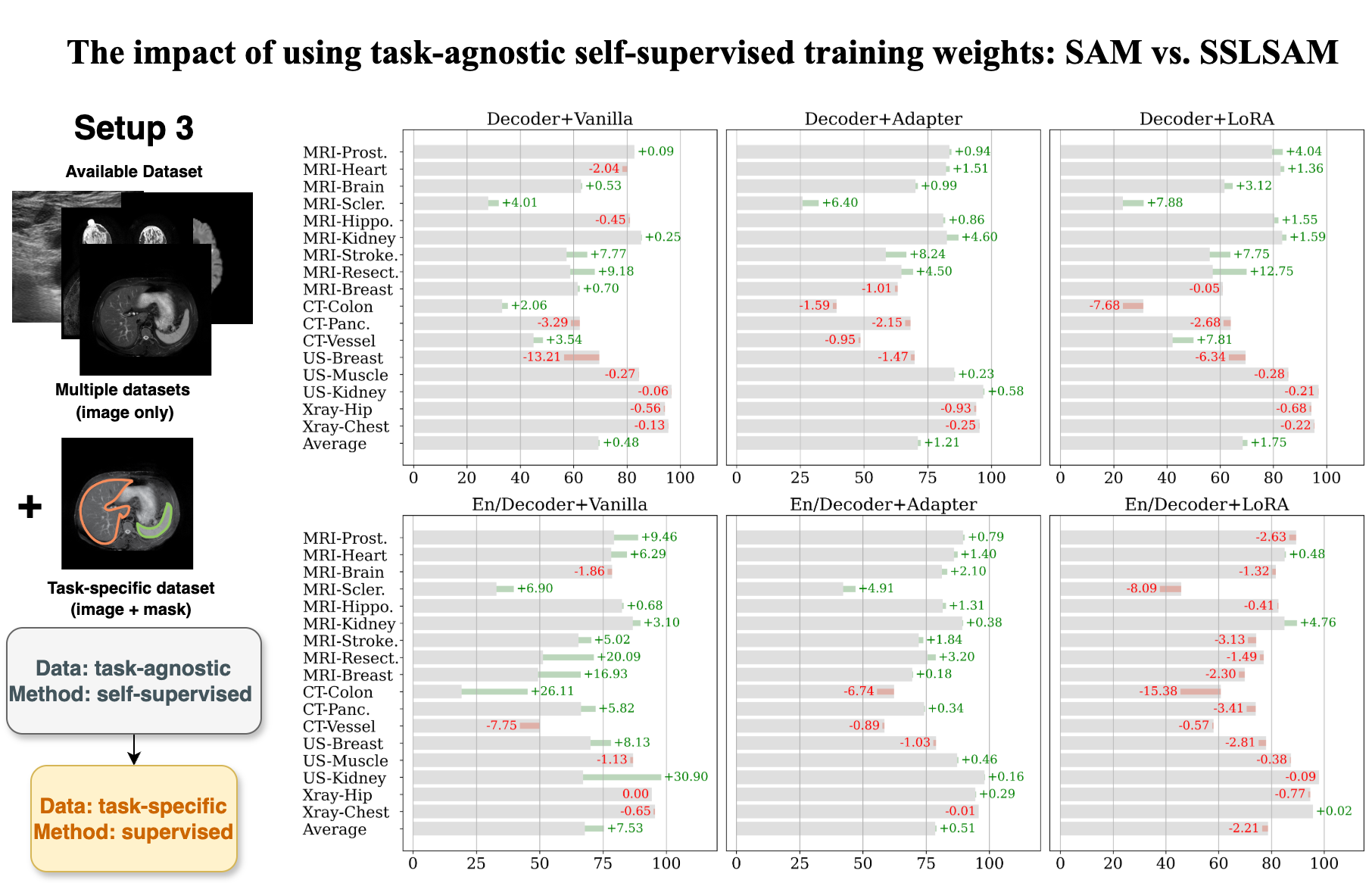}
    \caption{The comparison performance of applying various \stepone strategies to SAM {directly} (in gray) and the approach of using \stepthree first when additional labeled datasets are available. The differences are represented as SSLSAM - SAM, where green indicates a positive difference ($>$0) and red indicates a negative difference ($<$0), meaning that using SSLSAM to initiate fine-tuning led to a performance increase or drop, respectively.}
    
    \label{fig:ssl}
\end{figure*}

\subsubsection{Results}
The results are shown in Figure \ref{fig:sup} and Figure \ref{fig:medsam}, in which we list the performance of the original method when starting from SAM's initial weights in gray. We further illustrate the difference, highlighted in green (when positive) or red (if negative), when changing the initial weights from the original SAM to those acquired from \steptwo or MedSAM, respectively. 
When \stepone is applied solely to the decoder (the upper 3 blocks in both figures), we observe a consistent performance decrease in both scenarios.
Since in this case, phase 2 fine-tuning is applied exclusively to the decoder, it appears that weights learning through supervised training fail to adequately prepare the image encoder with the medical imaging domain knowledge necessary for automated feature extraction. 
Thus, attempting to conserve computational resources while starting with a publicly available, prompt-based medical model, such as MedSAM, becomes unnecessary if the goal is to update only the small decoder for customized datasets.

When fine-tuning both the encoder and decoder during phase 2 \stepone (Figure \ref{fig:sup} and \ref{fig:medsam}, bottom 3 blocks), we observe either a negative or no effect when utilizing a different pre-trained weights when using PEFT methods and a positive effect when using Vanilla fine-tuning. 
However, the latter setting leads to worse overall performance and thus offers little practical value.
Despite the decrement being alleviated when using MedSAM, the overall trend shows the ineffectiveness of prompt-based learning when the ultimate goal is automated segmentation, as MedSAM has a training set around 34 times larger yet results in no performance change at best.
We infer that the observed performance enhancement in En/Decoder-Vanilla is attributed to supervised pre-training, potentially nudging the weights closer to the medical domain, thereby reducing the likelihood of model collapse. 
Nonetheless, since this configuration still yields the lowest segmentation performance compared to the PEFT setting, this improvement does not offer a practical benefit.

Our research indicates that supervised pre-training across various labeled medical images, despite working well in the original setting (MedSAM ~\citep{MedSAM} is verified in its paper, and ours is verified in Section \ref{sec:scp_experiment}), fails to enhance automated segmentation performance. 
We believe the ineffectiveness of additional supervised pre-training is due to the difference in the tasks:
(1) The weights learned from interactive (prompt-based) mode vs. automated mode might be different and could not communicate with each other effectively. Additionally, extended pre-training may disrupt the transferability of the original weights.
(2) Furthermore, training the network on multiple segmentation tasks simultaneously could lead to conflicts within the training process, thus making the training more challenging. 
{These observations suggest that hybrid training, \textit{i.e.,} fine-tuning SAM for both the automatic segmentation task and the prompt-based segmentation task simultaneously, may not be effective.}

\subsection{Is \stepthree helpful for further automated segmentation?}
As discussed in Section \ref{section:self-supervised}, we select masked autoencoder (MAE) as the method for \stepthree. In this section, we provide the implementation details and analyze its performance.

\subsubsection{Implementation details}
In this stage, all MRI datasets in Table \ref{tab:data} were used during training.
The mask information was ignored, and only the images from the training set were utilized.
Blank slices were further removed since they were not informative for the reconstruction task.
This resulted in $159,104$ images for training.

Following the previous section, we only select ViT-B as the encoder architecture. To employ MAE's objective, we introduced an additional decoder to reconstruct the inputs, which was discarded after training. 
The additional decoder's structure, mask ratio, and learning rate schedule all follow that of the original paper.
Since the shape of SAM's input was restricted to $1024\times 1024$, we set the batch size to $2$.
To accommodate for the small batch size, we set the cumulative iteration, \ie the number of iterations to only compute gradients without updating the network parameters, to $16$. The training was conducted on 8 A6000 for $50$ epochs.

In this experimental setup, our methodology follows a similar two-phase procedure: (1) Phase 1: \stepthree. (2) Phase 2: \stepone for each dataset, beginning from pre-trained weights.
We name the weights after phase 1 as \textbf{SSLSAM}, which is shared publicly at https://github.com/mazurowski-lab/finetune-SAM.

\subsubsection{Results}
Figure \ref{fig:ssl} presents the results of using SSLSAM as the weights before \stepone.
We observe an improvement in the average performance of all MRIs across all architecture combinations besides En/Decoder + LoRA.
Noticeably, when selecting En/Decoder + Adapter as the fine-tuning strategy, we observe consistent improvements over all MRI datasets, and the model achieves an average performance of $78.09\%$, which surpasses the leading ViT-H based method by $1.08\%$ and the best UNet-based method (nnUNet) by $3.58\%$. 
Contradictory to \steptwo, the features learned during SSL are task agnostic, and since we only pre-train the encoder, there is no dependence on the task condition, which we have shown to hurt the performance.
In the non-MRI data sets base, we observe a minor decrease in performance. This may be due to the MRI's specificity.

\begin{figure*}[t]
    \centering
    \includegraphics[width=\textwidth]{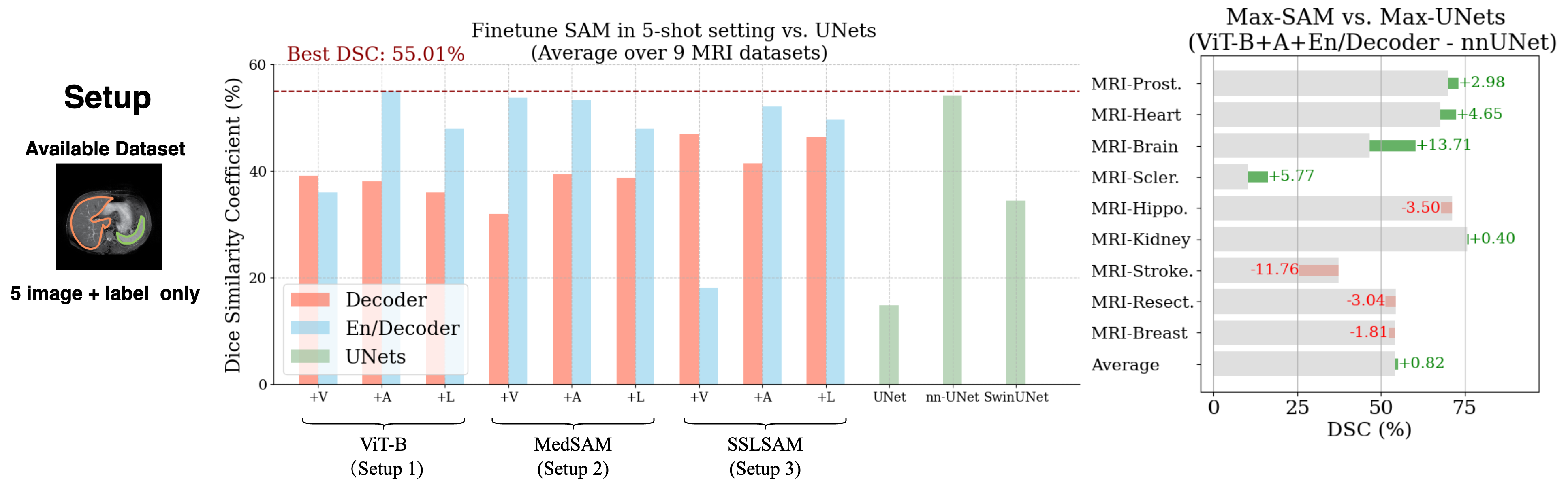}
    \caption{Left: Visualization of the performance of fine-tuned models compared with UNets in 5-shot setting (Section \ref{sec:fewshot}). Right: Difference between the best fine-tuned SAM model (ViT-B-A-En/Decoder) and the best UNet-based model (nnUnet) across individual datasets.}
    \label{fig:task-fewshot}
\end{figure*}

\subsection{Which fine-tuning setup works the best in the few-shot learning setting?}
\label{sec:fewshot}
\subsubsection{Implementation details} 
One of the popular scenarios for adopting fine-tuning strategies is when the target dataset has a limited number of annotated images, known as few-shot fine-tuning. Few-shot fine-tuning gets even more attention in the field of medical imaging segmentation, given the difficulty in securing a substantial quantity of expert annotations. We explore the effectiveness of different fine-tuning procedures in the following setting. 

Our study primarily investigates 5-shot learning, where we randomly select five images and their corresponding annotations from the training set and validation set, respectively. In the case of datasets derived from 3D volumes, these samples were chosen from distinct volumes to ensure diversity. 
In addition to the constraint of using only 5 training and validation images, we also modified the training hyperparameters to suit the few-shot setting better. 
Specifically, we set a maximum of 1000 epochs for training and a batch size of 2, and did not include any early stopping criteria. We increased the number of epochs to ensure sufficient updates for learning medical image patterns, used a smaller batch size to have diverse updates, and removed early stopping to prevent premature termination in the few-shot setting. We chose this strategy based on preliminary results indicating better performance compared to the default setting for full-set fine-tuning.

In this setting, we also investigated the previous three setups within a few-shot framework, including (1) task-specific fine-tuning exclusively and (2) two pre-trained setups: \steptwo and \stepthree, each followed by additional task-specific fine-tuning. 
Notably, we cannot use the weights learned from \steptwo as its training pipeline requires all labeled images. Thus, we consider the MedSAM weights as an alternative. 
In this section, we investigate three key questions: (1) whether additional pre-training offers superior initial weights for the few-shot setting, (2) which fine-tuning strategy proves most effective in a few-shot context, (3) whether fine-tuning foundation models still works better than UNets in the few-shot setting, and (4) how does performance vary with different levels of few-shot data availability, such as 10-shot or 20-shot?

\subsubsection{Results}
Figure \ref{fig:task-fewshot} shows the results for the 5-shot setting, comparing the fine-tuning of SAM's models with direct training on UNets.
The results demonstrate that using ViT-B + En/Decoder + Adapter on the initial SAM yields optimal performance across all configurations.
Comparing the results with UNets', especially with the best-performing nnUNet, fine-tuned SAM still manages to achieve a slight improvement. This observation aligns with the findings presented in Section \ref{section:task-specfic}. The performance of three selected datasets with different levels of few-shot data availability is shown in Appendix Figure \ref{fig:few-shot-diff-data}.

Consistently, the observation that MedSAM is ineffective persists in this few-shot scenario. 
While SSLSAM has shown effectiveness in the full-dataset setting, its efficacy diminished and even became detrimental in the few-shot scenario.
This observation aligns with the understanding that the task-agnostic training disrupts the model's connection with the segmentation tasks. While beneficial as initial weights for full-dataset training, this approach may not adequately rebuild these connections with only five images, resulting in lower performance. 
In summary, we observe the ineffectiveness of additional pre-training (either using \steptwo or \stepthree) with limited training data.
This can be contradictory to the common belief that imposing medical knowledge on the network in advance will be beneficial in the few-shot learning scenario.

\begin{figure*}[t]
    \centering
    \includegraphics[width=0.8\textwidth]{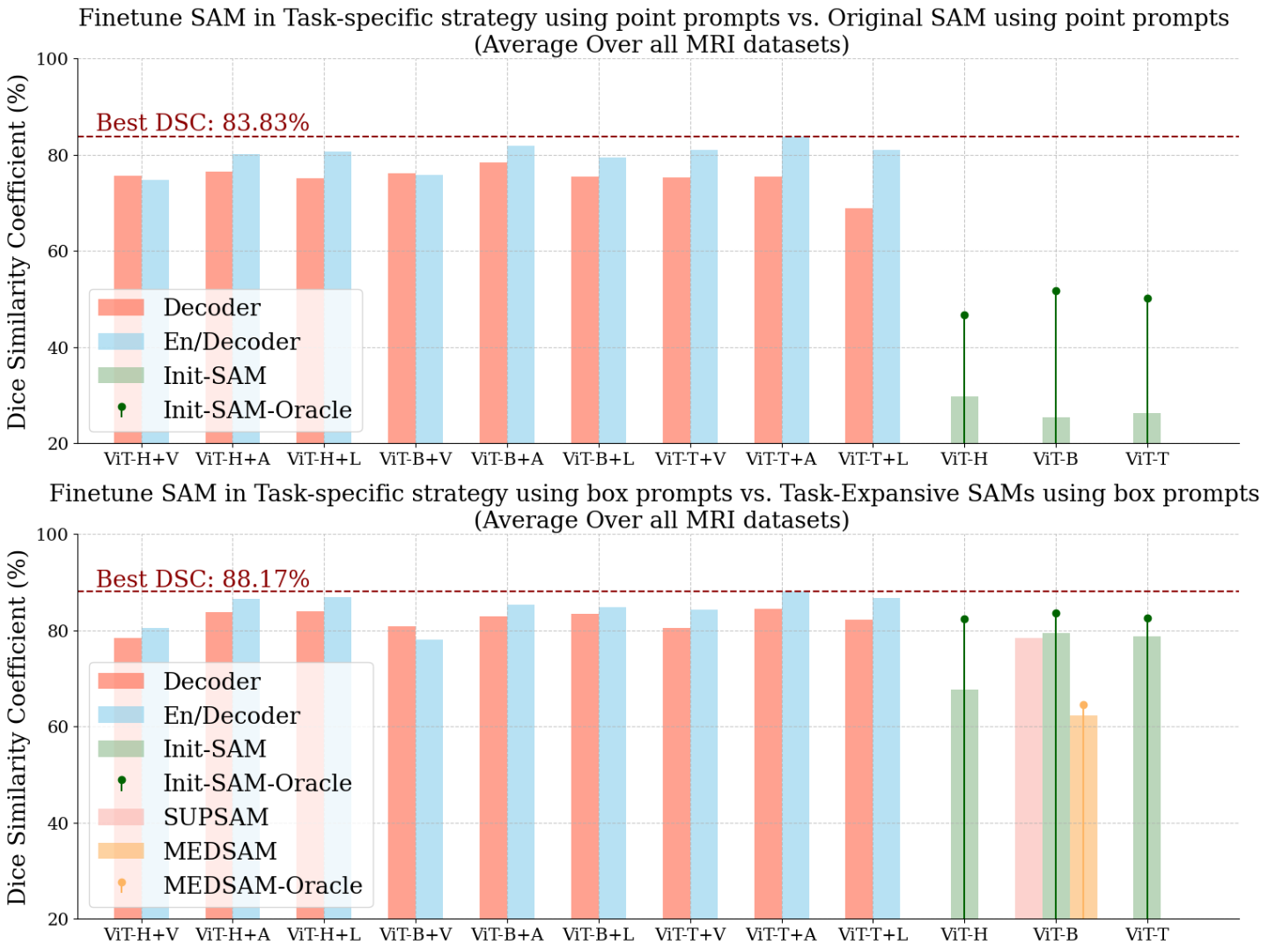}
    \caption{Performance of the task-specific fine-tuning strategies and initial SAM when using point/box prompts.}
    \label{fig:prompt-summary}
\end{figure*}

\begin{table*}[t]
    \centering
    \begin{tabular}{c|c|c}
       Dataset Setting & Condition & Optimal Strategy \\
       \hline
       Single Dataset    & NA & ViT-H + En/Decoder + LoRA \\
       Multiple Datasets & NA & SSLSAM + En/Decoder + Adapter \\
       SIngle Dataset & Few-shot & VIT-B + En/Decoder + Adapter \\
       Single Dataset & Prompt-based & ViT-T + En/Decoder + Adapter \\
       
    \end{tabular}
    \caption{{The optimal fine-tuning strategy under each scenario. SSLSAM refers to the weights obtained after self-supervised learning, introduced in Figure 1, Step 3. LoRA and Adapter are two parameter-efficient fine-tuning strategies, and En/Decoder means the adaption layers in both the encoder and decoder will be updated.}}
    \label{tab:my_label}
\end{table*}

\subsection{What if I can provide prompts to SAM?}
Despite the primary goal of this paper being to find the best practice to build an automated segmentation algorithm from the foundation models, we conduct a non-trivial extension that studies the effectiveness of various \stepone strategies in the interactive segmentation setting. 

\subsubsection{Implementation details}
We considered two types of prompts: point or box.
To generate a prompt, we utilized the same strategy as in \steptwo. 
The generated prompts were fed to the prompt encoder to obtain the prompt embeddings during both training and inference. The prompt encoder was fixed.
We also limited the evaluation data to MRI only since this task was not our main focus.

We included the performance of applying prompts to the original SAM and its oracle performance; the latter reflects the hypothetical performance when an extra step is taken to determine the best prediction from the network.
Additionally, we include weights acquired from \steptwo and MedSAM in the box-prompt setting.
When applying \stepone to SAM, we followed Section \ref{section:task-specfic} to include all strategies.

\subsubsection{Results}
\label{sec:prompt_result}
Figure \ref{fig:prompt-summary} shows the performance of all \stepone strategies in the interactive mode, as well as that of the original SAM as a reference. 
First, we observe that the performance of SAM with point prompts is not satisfactory. Replacing the points with boxes significantly improves performance. 
In this setting, we also examine the weights acquired from \steptwo and observe a similar performance, with a difference of only $0.92\%$ in DSC compared to the initial SAM. This observation confirms that the inefficiency of \steptwo is inherently due to having different tasks in mind.

To be noted, the best automated \stepone strategy yields an average performance of $77.01\%$ on all MRI datasets, which is better than the original SAM with point prompts and on par with that with box prompts.

Conducting \stepone is also shown to be effective in the interactive segmentation scenario, especially when points are used as the prompts. Moreover, we find that the best \stepone strategy is ViT-T + En/Decoder + Adapter in both scenarios, achieving $83.83\% / 88.17\%$ DSC when using a point/box as a prompt. This difference also suggests that providing box prompts can lead to better segmentation performance with the additional annotation cost.

\section{Conclusion}
\label{sec:conclusion}
In summary, we systematically study different strategies to fine-tune the Segment Anything Model (SAM) in the medical imaging field.
Our extensive experiments reveal several interesting observations and provide a guide for developing an automated segmentation algorithm from SAM.
Our findings are based on dataset availability and are summarized as follows:

When only a single dataset is available for fine-tuning, we find that:
\begin{enumerate}
    \item Fine-tuning SAM, if with proper strategies, leads to better segmentation performance than training from scratch with nnUNet ~\citep{isensee2021nnu}. 
    \item Increasing the size of network architecture leads to a marginal increase in segmentation performance, while exponentially longer training time.
    \item Using parameter-efficient fine-tuning (PEFT) techniques has a clear advantage over fine-tuning all parameters on medium-to-large-sized architectures (ViT-H or ViT-B), but not on small architectures (ViT-T).
    \item Optimizing both the encoder and decoder (En/De) has clear advantages over only the decoder, with the latter being a popular choice in the literature.
    \item In summary, we recommend using \textbf{ViT-B + En/Decoder + PEFT} as the fine-tuning framework because of its leading performance and moderate computation cost.
\end{enumerate}

When having multiple datasets, we can incorporate general medical knowledge into SAM through additional pre-training, and find that:
\begin{enumerate}
    \item Using additional labeled data for other tasks with prompt-based learning is an inefficient strategy to further improve SAM's performance.
    \item Using additional unlabeled data with self-supervised learning can improve SAM's performance when the test data is from the same modality as the pre-trained data, and has no effect otherwise.
\end{enumerate}

When we adjust the fine-tuning task into few-shot learning, we find that:
\begin{enumerate}
    \item Fine-tuning SAM significantly outperforms UNet, and still maintains a slight improvement compared with nn-UNet in the 5-shot setting.
    \item Additional \steptwo or \stepthree do not contribute to improvements in the final few-shot learning task.
\end{enumerate}

Lastly, in the interactive segmentation scenario, we observe that:
\begin{enumerate}
    \item Fine-tuned SAM brings substantial performance improvements compared to SAM and MedSAM in the interactive segmentation setting.
    \item Our recommendation framework is \textbf{ViT-T + En/Decoder + PEFT} given its superior performance and low computation cost.
\end{enumerate}

In addition to these findings, we also observe several limitations when fine-tuning SAM. 
Namely, the computational cost of fine-tuning SAM can be high when using the default backbones. As shown in Table \ref{tab:param_table}, when using ViT-H or ViT-B as the backbone, the number of parameters is 640.95M and 93.60 M. Using parameter-efficient fine-tuning strategies leads to better segmentation performance but does not improve the training efficiency. A potential solution is to utilize smaller backbones, such as ViT-T, \textit{i.e.,} a lightweight encoder distilled from pre-trained ViT-H ~\citep{zhang2023towards}, to balance the performance and computation cost. \\
Another limitation is that our study only focuses on the 2D segmentation tasks. 
To alleviate this issue, we have conducted a preliminary study that utilizes nnUNet 3D as a strong 3D segmentation baseline. We observed that the performance of DSC(\%) is 87.14 vs. 87.35, 40.56 vs. 29.96, 47.58 vs. 46.95 on MRI-Heart, MRI-Scler., and CT-Colon, respectively, when comparing between nnUNet 3D vs. 2D. This observation suggests that the improvements are limited when the baseline methods are switched from 2D to 3D.
On the other hand, adapting SAM to 3D segmentation is beyond the scope of this paper. We believe some potential directions include adapting SAM directly to 3D ~\citep{gong20243dsam} or fine-tuning the recently proposed SAM 2 ~\citep{ravi2024sam}, which is designed specifically for video segmentation.


\acks{Research reported in this publication was supported by the National Institute Of Biomedical Imaging And Bioengineering of the National Institutes of Health under Award Number R01EB031575. The content is solely the responsibility of the authors and does not necessarily represent the official views of the National Institutes of Health.}

%
\ethics{The work follows appropriate ethical standards in conducting research and writing the manuscript, following all applicable laws and regulations regarding treatment of animals or human subjects.}

\coi{The conflicts of interest have not been entered yet.}

\data{All data used in this study are publicly available. We provide detailed descriptions and proper citations for each dataset used to ensure full transparency and reproducibility. Additionally, all code for data preprocessing and preparation is publicly released and can be used to fully reproduce our results. The code is available at: \url{https://github.com/mazurowski-lab/finetune-SAM}.}

\bibliography{sample}


\clearpage
\appendix
\section{Proof of the central theorem}
	In this appendix we prove the central theorem and present additional experimental results.
	\noindent

	{\noindent \em Remainder omitted in this sample. }

\section{Appendix section}
In the appendix, we provide the numerical results of all experiments. 

\begin{figure*}[h!]
    \centering
    \includegraphics[width=0.6\linewidth]{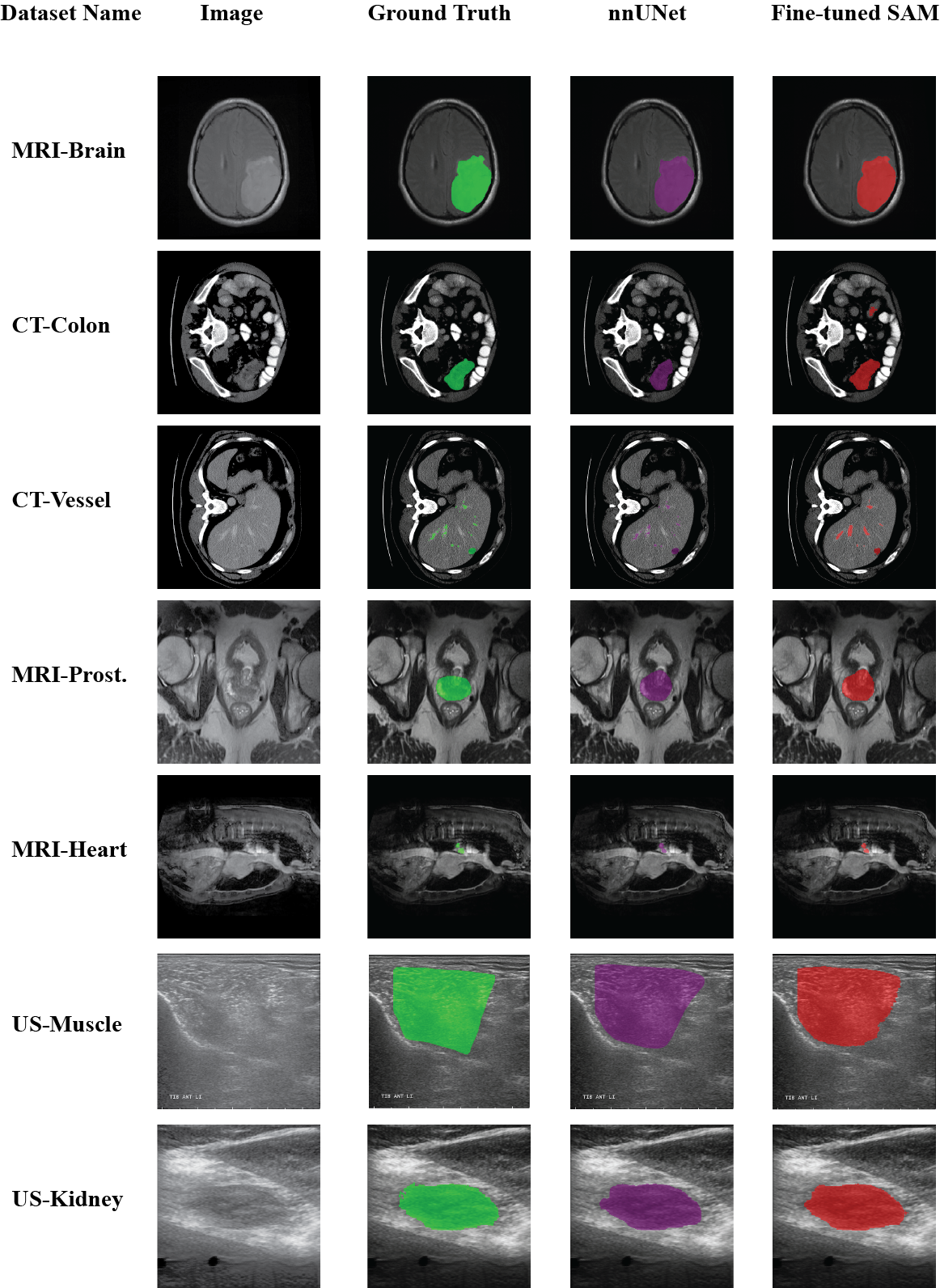}
    \caption{{Comparison of Segmentation Results: Examples showcasing the performance of the foundation model (ViT-H + L + Encoder/Decoder) versus the non-foundation-based method (nn-UNet).}}
    \label{fig:sam_vs_nnunet}
\end{figure*}

Tables \ref{tab:ap_single_vith}, \ref{tab:ap_single_vitb}, \ref{tab:ap_single_vitt} present the results of applying \stepone to ViT-H, ViT-B, and ViT-T respectively, and table \ref{tab:ap_unet} shows the performance of UNets. Figure \ref{fig:task-specific} summarizes the results of these 4 tables, {and Figure \ref{fig:sam_vs_nnunet} visualizes some examples of segmentation results of using foundation model vs. using nnUNet.}

Tables \ref{tab:ap_sup}, \ref{tab:ap_med}, \ref{tab:ap_ssl} show the performance of replacing SAM's weights with SupSAM (acquired from \steptwo), MedSAM (from \cite{MedSAM}), and SSLSAM (acquired from \stepthree), respectively. Since all weights utilize the same encoder architecture (ViT-B), we show the difference by conducting Table \ref{tab:ap_single_vitb} minus the corresponding table in Figure \ref{fig:sup}, \ref{fig:medsam}, and \ref{fig:ssl}, respectively.

\begin{figure*}[h!]
    \centering
    \includegraphics[width=0.9\linewidth]{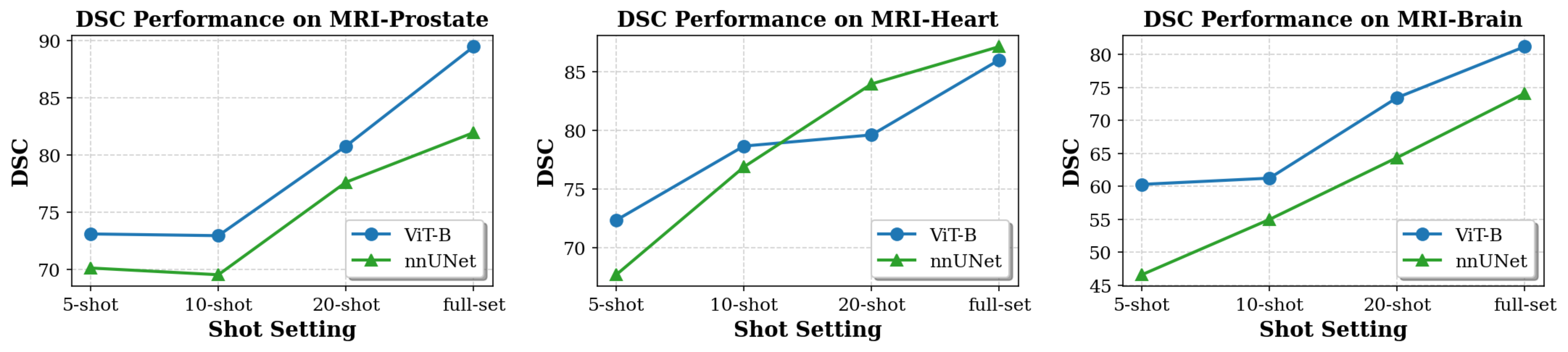}
    \caption{{Visuallization of three selected datasets' model (ViT-B, EN/Decoder, Adapter) performance (DSC) vs. different dataset availability: 5-shot, 10-shot, 20-shot, and the full set.} }
    \label{fig:few-shot-diff-data}
\end{figure*}

Tables \ref{tab:ap_fs_unet} and \ref{tab:ap_fs_sam} show the performance of applying \stepone in the few-shot learning setting, which is summarized in Figure \ref{fig:task-fewshot}. The results for the selected three datasets of expanded few-shot learning: 10-shot, 20-shot, are shown in Figure \ref{fig:few-shot-diff-data}.

Table \ref{tab:ap_point_ori} shows the performance of initial SAM in the point-based segmentation setting, with tables \ref{tab:ap_pointh}, \ref{tab:ap_pointb}, and \ref{tab:ap_pointt} showing applying \stepone in the same setting. Tables \ref{tab:ap_box_ori}, \ref{tab:ap_boxh}, \ref{tab:ap_boxb}, and \ref{tab:ap_boxt} are conducted in the same setting but with the box as the prompt. All results are summarized in Figure \ref{fig:prompt-summary}.

\begin{table*}[h]
    \centering
    \footnotesize
    \setlength{\tabcolsep}{4pt}
    \begin{tabular}{l|c|c|c|c|c|c|c|c|c|c|c|c|c|c}
        \multirow{4}{*}{Dataset} & \multicolumn{14}{c}{ViT-H} \\ \cline{2-15}
        & \multicolumn{6}{c|}{Decoder} & \multicolumn{8}{c}{En/Decoder} \\ \cline{2-15}
        & \multicolumn{2}{c|}{V}  & \multicolumn{2}{c|}{A} & \multicolumn{2}{c|}{L} & \multicolumn{2}{c|}{V} & \multicolumn{2}{c|}{A} & \multicolumn{2}{c|}{A*} & \multicolumn{2}{c}{L} \\ \cline{2-15}
        & DSC  & {NSD} & DSC  & {NSD} & DSC  & {NSD} & DSC  & {NSD} & DSC  & {NSD} & DSC  & {NSD} & DSC  & {NSD}\\
        \hline
        MRI-Prost. & $82.58$  & {$31.37$} & $82.42$ & {$33.33$} & $78.60$ & {$35.08$} & $73.79$ & {$18.47$} & $88.97$ & {$42.61$} & $88.80$ & {$47.09$} & $89.64$ & {$50.01$} \\
        MRI-Heart  & $80.86$  & {$64.16$} & $80.61$ & {$62.69$} & $80.53$ & {$58.43$} & $73.69$ & {$48.07$} & $85.07$ & {$75.47$} & $86.22$ & {$76.76$} & $85.07$ & {$74.35$} \\
        MRI-Brain  & $45.32$ & {$19.37$}  & $60.19$ & {$26.92$} & $50.87$ & {$23.94$} & $53.89$ & {$15.06$} & $80.53$ & {$42.41$} & $82.65$ & {$54.43$} & $84.09$ & {$62.38$} \\
        MRI-Scler. & $21.16$ & {$21.03$}  & $22.01$ & {$21.18$} & $22.51$ & {$11.04$} & $27.81$ & {$22.36$} & $42.74$ & {$20.04$} & $48.44$ & {$48.32$} & $47.24$ & {$47.96$} \\
        MRI-Hippo. & $81.25$ & {$29.43$}  & $80.30$ & {$26.83$} & $79.01$ & {$22.84$} & $81.22$ & {$27.22$} & $81.61$ & {$20.61$} & $82.56$ & {$29.08$} & $83.33$ & {$43.17$} \\
        MRI-Kidney & $82.34$ & {$57.39$}  & $84.08$ & {$58.66$} & $80.60$ & {$48.60$} & $87.65$ & {$62.25$} & $88.81$ & {$68.05$} & $90.29$ & {$68.47$} & $83.02$ & {$58.77$} \\
        MRI-Stroke.& $47.52$ & {$35.99$}  & $53.06$ & {$40.32$} & $47.22$ & {$34.59$} & $56.51$ & {$52.79$} & $71.64$ & {$60.21$} & $67.82$ & {$55.95$} & $72.39$ & {$60.86$} \\
        MRI-Resect.& $43.96$ & {$26.17$}    & $60.09$ & {$42.77$} & $54.23$ & {$31.38$} & $57.26$ & {$38.33$} & $75.03$ & {$61.56$} & $67.77$ & {$35.28$} & $76.62$ & {$65.12$} \\
        MRI-Breast & $59.48$ & {$66.62$}    & $61.10$ & {$67.89$} & $59.05$ & {$65.72$} & $60.41$ & {$66.50$} & $69.00$ & {$74.91$} & $69.11$ & {$74.05$} & $71.71$ & {$77.69$} \\
        \hline
        CT-Colon  & $22.67$ & {$17.00$}     & $35.08$ & {$26.53$} & $26.77$ & {$17.40$} & $24.12$ & {$12.65$} & $55.44$ & {$42.78$} & $54.30$ & {$43.06$} & $60.91$ & {$55.61$} \\
        CT-Panc.  & $61.78$ & {$41.37$}     & $67.43$ & {$46.85$} & $64.06$ & {$43.96$} & $64.45$ & {$40.67$} & $73.92$ & {$55.41$} & $73.01$ & {$53.38$} & $75.74$ & {$59.90$} \\
        CT-vessel.& $40.94$ & {$45.68$}     & $44.75$ & {$49.84$} & $42.15$ & {$46.23$} & $40.96$ & {$54.58$} & $58.16$ & {$55.81$} & $54.67$ & {$50.22$} & $58.53$ & {$60.53$} \\
        US-Breast & $71.27$ & {$26.74$}  & $74.17$ & {$28.43$} & $70.22$ & {$25.86$} & $64.91$ & {$18.29$} & $77.33$ & {$32.43$} & $78.18$ & {$33.09$} & $77.36$ & {$36.41$} \\
        US-Muscle & $84.50$ & {$19.28$}  & $85.52$ & {$20.39$} & $85.05$ & {$20.50$} & $86.66$ & {$23.12$} & $86.16$ & {$19.29$} & $86.92$ & {$20.15$} & $87.49$ & {$24.13$} \\
        US-Kidney & $96.97$ & {$68.83$}   & $97.08$ & {$69.05$}& $96.53$ & {$62.97$} & $71.34$ & {$21.92$} & $98.13$ & {$81.29$} & $98.34$ & {$82.05$} & $98.18$ & {$81.88$} \\
        \hline
        Xray-Hip  & $93.78$ & {$73.71$}   & $94.35$ & {$72.17$}& $93.00$ & {$74.47$} & $92.02$ & {$70.84$} & $93.96$ & {$75.77$} & $94.33$ & {$73.78$} & $94.73$ & {$73.44$} \\
        Xray-Chest& $95.29$ & {$63.29$}  & $95.35$ & {$64.03$}& $95.22$ & {$62.69$} & $94.85$ & {$56.99$} & $95.81$ & {$69.49$} & $95.45$ & {$70.11$} & $95.74$ & {$70.56$} \\
        \hline
        \hline
        Average   & $65.36$ & {$41.86$} & $69.24$ & {$44.81$} & $66.34$ & {$40.57$} & $65.39$ & {$38.24$} & $77.35$ & {$52.83$} & $77.58$ & {$53.84$} & $78.93$ & {$58.99$} \\
    \end{tabular}
    \caption{The performance of SAM under different fine-tuning strategies in the automated mode across different datasets. V = vanilla, A = Adapter, L = LoRA. A* is a variant of A in which we applied Adapter after each network layer.}
    \label{tab:ap_single_vith}
\end{table*}

\begin{table*}[h]
    \centering
    \footnotesize
    \begin{tabular}{l|c|c|c|c|c|c|c|c|c|c|c|c}
        \multirow{4}{*}{Dataset} & \multicolumn{12}{c}{ViT-B} \\ \cline{2-13}
        & \multicolumn{6}{c|}{Decoder} & \multicolumn{6}{c}{En/Decoder} \\ \cline{2-13}
        & \multicolumn{2}{c|}{V}  & \multicolumn{2}{c|}{A} & \multicolumn{2}{c|}{L} & \multicolumn{2}{c|}{V} & \multicolumn{2}{c|}{A} & \multicolumn{2}{c}{L} \\ \cline{2-13}
        & DSC  & {NSD} & DSC  & {NSD} & DSC  & {NSD} & DSC  & {NSD} & DSC  & {NSD} & DSC  & {NSD}\\
        \hline
        MRI-Prost. 
        & $82.77$ & {$33.71$} & $83.44$ & {$36.50$} & $79.51$ & {$29.78$} & $79.36$ & {$25.69$} & $89.50$ & {$49.88$} & $89.37$ & {$42.66$} \\
        MRI-Heart 
        & $80.29$ & {$63.53$} & $82.16$ & {$66.50$} & $80.90$ & {$62.60$} & $78.16$ & {$51.79$} & $86.01$ & {$76.92$} & $85.04$ & {$24.47$} \\
        MRI-Brain 
        & $62.73$ & {$29.46$} & $70.23$ & {$36.56$} & $65.35$ & {$31.54$} & $78.57$ & {$42.86$} & $81.21$ & {$47.26$} & $81.65$ & {$57.61$} \\
        MRI-Scler. 
        & $27.95$ & {$28.34$} & $25.92$ & {$25.54$} & $27.91$ & {$26.47$} & $32.90$ & {$28.67$} & $42.12$ & {$37.57$} & $45.83$ & {$32.28$} \\
        MRI-Hippo.  
        & $81.08$ & {$31.21$} & $81.09$ & {$27.50$} & $80.35$ & {$30.07$} & $82.45$ & {$28.69$} & $81.52$ & {$28.67$} & $82.69$ & {$26.36$} \\
        MRI-Kidney  
        & $85.29$ & {$60.88$} & $82.50$ & {$53.40$} & $79.58$ & {$42.45$} & $86.70$ & {$59.03$} & $89.19$ & {$68.82$} & $84.93$ & {$72.13$} \\
        MRI-Stroke. 
        & $57.33$ & {$43.97$} & $58.47$ & {$45.58$} & $52.67$ & {$35.47$} & $65.32$ & {$55.95$} & $72.05$ & {$61.61$} & $74.24$ & {$63.65$} \\
        MRI-Resect.
        & $58.74$ & {$39.84$} & $64.76$ & {$48.44$} & $57.23$ & {$7.85$} & $51.28$ & {$22.87$} & $75.53$ & {$63.25$} & $71.23$ & {$65.18$} \\
        MRI-Breast  
        & $61.60$ & {$68.69$} & $63.16$ & {$70.46$} & $61.02$ & {$68.56$} & $49.30$ & {$64.10$} & $69.54$ & {$74.92$} & $69.95$ & {$76.68$} \\
        \hline
        CT-Colon   
        & $33.25$ & {$22.30$} & $39.29$ & {$30.17$} & $31.12$ & {$21.64$} & $19.18$ & {$11.77$} & $62.30$ & {$54.53$} & $60.87$ & {$53.95$} \\
        CT-Panc. 
        & $62.28$ & {$38.55$} & $68.34$ & {$48.62$} & $63.97$ & {$42.83$} & $66.25$ & {$45.00$} & $74.16$ & {$57.06$} & $74.07$ & {$57.58$} \\
        CT-vessel. 
        & $44.95$ & {$48.46$} & $48.71$ & {$51.39$} & $42.16$ & {$48.28$} & $49.89$ & {$53.71$} & $58.56$ & {$60.45$} & $58.08$ & {$59.97$} \\
        \hline
        US-Breast   
        & $69.64$ & {$27.40$} & $69.93$ & {$23.73$} & $69.56$ & {$24.00$} & $70.02$ & {$25.38$} & $78.87$ & {$34.87$} & $77.97$ & {$32.92$} \\
        US-Muscle   
        & $84.53$ & {$19.51$} & $85.53$ & {$19.19$} & $85.61$ & {$21.12$} & $86.80$ & {$23.43$} & $87.24$ & {$24.20$} & $87.37$ & {$23.90$} \\
        US-Kidney   
        & $96.70$ & {$66.50$} & $96.76$ & {$66.41$} & $96.99$ & {$67.93$} & $67.12$ & {$12.29$} & $98.08$ & {$80.80$} & $98.05$ & {$80.35$} \\
        \hline
        Xray-Hip  
        & $94.29$ & {$70.04$} & $94.11$ & {$68.23$} & $97.68$ & {$71.44$} & $94.14$ & {$71.53$} & $94.44$ & {$79.32$} & $94.73$ & {$81.01$} \\
        Xray-Chest 
        & $95.45$ & {$65.18$} & $95.29$ & {$62.85$} & $95.45$ & {$65.48$} & $95.58$ & {$66.31$} & $95.73$ & {$69.87$} & $95.78$ & {$69.47$} \\
        \hline
        \hline
        Average  
        & $69.80$ & {$44.81$} & $71.16$ & {$46.22$} & $64.77$ & {$40.11$} & $67.82$ & {$40.53$} & $78.59$ & {$57.05$} & $78.69$ & {$54.12$} \\
    \end{tabular}
    \caption{The performance of SAM under different fine-tuning strategies in the automated mode across different datasets. V = vanilla, A = Adapter, L = LoRA.}
    \label{tab:ap_single_vitb}
\end{table*}

\begin{table*}[h]
    \centering
    {\footnotesize
    \begin{tabular}{l|c|c|c|c|c|c|c|c|c|c|c|c}
        \multirow{4}{*}{Dataset} & \multicolumn{12}{c}{ViT-T} \\ \cline{2-13}
        & \multicolumn{6}{c|}{Decoder} & \multicolumn{6}{c}{En/Decoder} \\ \cline{2-13}
        & \multicolumn{2}{c|}{V}  & \multicolumn{2}{c|}{A} & \multicolumn{2}{c|}{L} & \multicolumn{2}{c|}{V} & \multicolumn{2}{c|}{A} & \multicolumn{2}{c}{L} \\ \cline{2-13}
        & DSC  & {NSD} & DSC  & {NSD} & DSC  & {NSD} & DSC  & {NSD} & DSC  & {NSD} & DSC  & {NSD}\\
        \hline
        MRI-Prost.  
        & $81.36$ & {$32.07$} & $80.01$ & {$30.26$} & $77.17$ & {$27.22$} & $88.90$ & {$49.55$} & $89.37$ & {$48.57$} & $86.70$ & {$41.30$} \\
        MRI-Heart  
        & $80.75$ & {$64.09$} & $80.39$ & {$60.28$} & $76.24$ & {$50.87$} & $85.15$ & {$74.76$} & $84.93$ & {$73.79$} & $84.35$ & {$72.54$} \\
        MRI-Brain  
        & $52.68$ & {$23.88$} & $58.43$ & {$26.55$} & $28.12$ & {$8.20$} & $78.32$ & {$53.18$} & $79.42$ & {$56.09$} & $77.14$ & {$52.36$} \\
        MRI-Scler. 
        & $21.20$ & {$20.38$} & $18.22$ & {$15.84$} & $15.29$ & {$14.47$} & $42.60$ & {$43.84$} & $41.05$ & {$40.98$} & $41.92$  & {$42.54$} \\
        MRI-Hippo.
        & $78.97$ & {$28.95$} & $78.39$ & {$23.73$} & $76.00$ & {$23.74$} & $82.90$ & {$44.45$} & $82.96$ & {$43.85$} & $82.56$ & {$39.60$} \\
        MRI-Kidney 
        & $82.49$ & {$52.68$} & $79.93$ & {$46.70$} & $67.54$ & {$36.20$} & $89.85$ & {$70.32$} & $88.39$ & {$68.55$} & $88.38$ & {$68.96$} \\
        MRI-Stroke. 
        & $42.13$ & {$30.31$} & $46.35$ & {$34.00$} & $34.08$ & {$19.68$} & $71.02$ & {$59.46$} & $72.14$ & {$60.31$} & $70.37$ & {$58.82$} \\
        MRI-Resect. 
        & $40.35$ & {$27.88$} & $40.56$ & {$21.47$} & $37.10$ & {$19.96$} & $74.34$ & {$61.28$} & $75.74$ & {$63,86$} & $72.93$ & {$60.12$} \\
        MRI-Breast 
        & $53.06$ & {$61.55$} & $55.10$ & {$62.58$} & $45.23$ & {$54.33$} & $66.92$ & {$74.75$} & $68.28$ & {$76.27$} & $66.18$ & {$72.84$} \\
        \hline
        CT-Colon  
        & $14.97$ & {$11.16$} & $23.09$ & {$14.35$} & $23.16$ & {$15.50$} & $55.99$ & {$48.99$} & $53.76$ & {$47.94$} & $49.63$ & {$41.11$} \\
        CT-Panc.   
        & $59.01$ & {$38.52$} & $64.54$ & {$42.06$} & $55.43$ & {$33.58$} & $72.08$ & {$54.74$} & $73.21$ & {$57.13$} & $70.36$ & {$52.21$} \\
        CT-vessel. 
        & $37.25$ & {$41.47$} & $36.55$ & {$37.00$} & $32.25$ & {$37.85$} & $56.62$ & {$59.39$} & $57.25$ & {$59.08$} & $55.09$ & {$56.54$} \\
        \hline
        US-Breast  
        & $61.42$ & {$18.59$} & $67.38$ & {$21.88$} & $49.64$ & {$13.28$} & $77.55$ & {$35.52$} & $74.50$ & {$32.32$} & $71.15$ & {$30.16$} \\
        US-Muscle  
        & $84.68$ & {$19.37$} & $85.50$ & {$19.87$} & $59.94$ & {$7.15$} & $86.72$ & {$23.67$} & $86.71$ & {$22.52$} & $87.18$ & {$23.94$} \\
        US-Kidney  
        & $96.11$ & {$61.66$} & $96.37$ & {$59.30$} & $92.60$ & {$44.55$} & $97.45$ & {$81.29$} & $98.06$ & {$80.41$} & $97.88$ & {$78.02$} \\
        \hline
        Xray-Hip    
        & $97.58$ & {$68.87$} & $93.83$ & {$64.80$} & $91.04$ & {$46.13$} & $92.33$ & {$76.82$} & $94.33$ & {$76.34$} & $94.61$ & {$76.76$} \\
        Xray-Chest
        & $95.08$ & {$60.87$} & $95.26$ & {$62.78$} & $91.50$ & {$43.07$} & $95.87$ & {$70.25$} & $95.75$ & {$68.62$} & $95.72$ & {$68.49$} \\
        \hline
        \hline
        Average    
        & $63.27$ & {$39.22$} & $64.57$ & {$38.10$} & $56.02$ & {$29.34$} & $77.33$ & {$58.11$} & $77.40$ & {$57.78$} & $76.01$ & {$55.36$} \\
    \end{tabular}
    }
    \caption{The performance of SAM under different fine-tuning strategies in the automated mode across different datasets. V = Vanilla, A = Adapter, L = LoRA.}
    \label{tab:ap_single_vitt}
\end{table*}

\begin{table*}[h]
    \centering
    \begin{tabular}{l|c|c|c|c|c|c}
        \multirow{2}{*}{Dataset} & \multicolumn{2}{c|}{UNet} & \multicolumn{2}{c|}{nnUNet} & \multicolumn{2}{c}{SwinUNet}\\ \cline{2-7}
        & DSC  & {NSD} & DSC  & {NSD} & DSC  & {NSD} \\
        \hline
        MRI-Prost.  & $82.59$ & {$11.64$} & $81.98$ & {$46.39$} & $82.03$ & {$10.78$} \\
        MRI-Heart   & $82.85$ & {$25.12$} & $87.14$ & {$79.51$} & $84.10$ & {$32.13$} \\
        MRI-Brain   & $79.00$ & {$16.97$} & $74.13$ & {$50.37$} & $75.83$ & {$18.20$} \\
        MRI-Scler.  & $36.71$  & {$12.38$} & $40.56$ & {$39.93$} & $36.73$ & {$11.69$} \\
        MRI-Hippo.  & $80.59$ & {$14.53$} & $76.18$ & {$37.62$} & $83.25$ & {$26.12$} \\
        MRI-Kidney  & $87.03$ & {$22.60$} & $91.69$ & {$72.90$} & $83.83$ & {$20.85$} \\
        MRI-Stroke. & $69.67$ & {$26.84$} & $72.27$ & {$55.86$} & $70.45$ & {$35.97$} \\
        MRI-Resect. & $71.07$ & {$27.88$} & $74.15$ & {$63.60$} & $65.91$ & {$21.41$} \\
        MRI-Breast  & $68.77$ & {$56.02$} & $72.49$ & {$75.52$} & $70.33$ & {$57.44$} \\
        \hline
        CT-Colon    & $43.64$ & {$12.04$} & $47.58$ & {$43.25$} & $46.39$ & {$12.00$} \\
        CT-Panc.    & $70.06$ & {$17.38$} & $74.78$ & {$58.07$} & $68.54$ & {$15.96$} \\ 
        CT-vessel.  & $57.06$ & {$31.49$} & $58.44$ & {$58.50$} & $56.06$ & {$31.89$} \\
        \hline
        US-Breast   & $64.83$ & {$5.26$} & $76.03$ & {$33.56$} & $72.02$ & {$7.80$} \\
        US-Muscle   & $83.99$ & {$4.99$} & $87.30$ & {$25.77$} & $86.62$ & {$7.32$} \\
        US-Kidney   & $96.74$ & {$43.96$} & $98.37$ & {$83.16$} & $97.75$ & {$51.09$} \\
        \hline   
        Xray-Hip    & $97.86$ & {$33.56$} & $98.46$ & {$83.52$} & $94.29$ & {$33.41$} \\
        Xray-Chest  & $95.51$ & {$23.82$} & $96.40$ & {$70.29$} & $95.62$ & {$23.89$} \\
        \hline
        \hline
        Average     & $74.44$ & {$22.86$} & $76.94$ & {$57.52$} & $74.69$ & {$24.71$}  
    \end{tabular}    \caption{The performance of UNet, nnUNet, and SwinUNet across different datasets.}
    \label{tab:ap_unet}
\end{table*}

\begin{table*}[h]
    \centering
    \footnotesize
   \begin{tabular}{l|c|c|c|c|c|c|c|c|c|c|c|c}
        \multirow{4}{*}{Dataset} & \multicolumn{12}{c}{SupSAM} \\ \cline{2-13}
        & \multicolumn{6}{c|}{Decoder} & \multicolumn{6}{c}{En/Decoder} \\ \cline{2-13}
        & \multicolumn{2}{c|}{V}  & \multicolumn{2}{c|}{A} & \multicolumn{2}{c|}{L} & \multicolumn{2}{c|}{V} & \multicolumn{2}{c|}{A} & \multicolumn{2}{c}{L} \\ \cline{2-13}
        & DSC  & {NSD} & DSC  & {NSD} & DSC  & {NSD} & DSC  & {NSD} & DSC  & {NSD} & DSC  & {NSD}\\
        \hline
        MRI-Prost. & $71.48$  & {$18.20$} & $72.12$  & {$19.67$} & $81.98$  & {$31.59$} & $82.01$  & {$29.76$} & $81.65$  & {$29.06$} & $83.63$ & {$31.45$} \\
        MRI-Heart & $79.78$  & {$58.22$} & $75.57$  & {$48.18$} & $76.34$  & {$51.57$} & $83.95$  & {$69.98$} & $82.16$  & {$65.93$} & $83.94$ & {$66.78$} \\
        MRI-Brain & $59.44$  & {$27.04$} & $63.61$  & {$29.37$} & $62.68$  & {$30.42$} & $79.95$  & {$42.81$} & $76.89$  & {$41.55$} & $75.07$ & {$39.52$} \\
        MRI-Scler. & $20.96$  & {$18.96$} & $22.10$  & {$19.53$} & $18.84$  & {$16.20$} & $34.32$  & {$29.70$} & $32.88$  & {$31.32$} & $32.58$ & {$31.24$} \\
        MRI-Hippo. & $80.91$  & {$27.00$} & $79.83$  & {$25.63$} & $79.98$  & {$26.10$} & $82.27$  & {$28.68$} & $81.98$  & {$29.73$} & $81.74$ & {$28.03$} \\
        MRI-Kidney & $77.88$  & {$46.81$} & $79.99$  & {$49.42$} & $77.10$  & {$46.21$} & $87.99$  & {$62.83$} & $85.71$  & {$59.58$} & $85.45$ & {$59.88$} \\
        MRI-Stroke. & $62.63$  & {$51.19$} & $64.45$  & {$53.61$} & $62.68$  & {$50.60$} & $67.42$  & {$59.07$} & $67.77$  & {$58.47$} & $67.29$ & {$56.92$} \\
        MRI-Resect. & $58.51$  & {$36.59$} & $59.60$  & {$38.20$} & $55.47$  & {$34.02$} & $69.20$  & {$50.09$} & $64.99$  & {$46.51$} & $67.07$ & {$46.93$} \\
        MRI-Breast & $55.50$  & {$63.82$} & $57.57$  & {$65.76$} & $55.74$  & {$63.65$} & $58.06$  & {$66.64$} & $58.01$  & {$67.31$} & $58.19$ & {$67.81$} \\
        \hline
        CT-Colon & $15.07$  & {$10.17$} & $16.07$  & {$10.24$} & $18.39$  & {$10.70$} & $38.30$  & {$25.62$} & $33.68$  & {$24.19$} & $35.86$ & {$26.17$} \\
        CT-Panc.  & $47.16$  & {$23.66$} & $52.58$  & {$27.69$} & $45.22$  & {$23.81$} & $66.34$  & {$43.13$} & $62.56$  & {$39.01$} & $63.41$ & {$41.59$} \\
        CT-Vessel & $37.08$  & {$48.31$} & $40.87$  & {$50.24$} & $34.23$  & {$43.51$} & $50.27$  & {$52.09$} & $51.73$  & {$54.59$} & $51.67$ & {$54.21$} \\
        \hline
        US-Breast & $43.41$  & {$8.53$} & $47.48$  & {$8.89$} & $39.77$  & {$7.50$} & $73.44$  & {$28.24$} & $66.99$  & {$21.35$} & $68.15$ & {$21.98$} \\
        US-Muscle & $83.26$  & {$16.17$} & $82.18$  & {$15.89$} & $80.08$  & {$13.26$} & $87.05$  & {$23.04$} & $85.81$  & {$19.11$} & $86.27$ & {$20.46$} \\
        US-Kidney & $95.86$  & {$55.78$} & $91.83$  & {$43.45$} & $88.28$  & {$25.50$} & $97.69$ & {$74.37$}  & $96.61$  & {$65.56$} & $96.81$ & {$66.91$} \\
        \hline
        Xray-Hip & $90.90$  & {$38.34$} & $89.89$  & {$45.61$} & $90.12$  & {$36.53$} & $95.64$  & {$41.32$} & $93.03$  & {$61.71$} & $96.32$ & {$66.39$} \\
        Xray-Chest & $93.56$  & {$48.37$} & $93.46$  & {$47.15$} & $92.25$  & {$41.28$} & $94.95$  & {$62.11$} & $95.28$  & {$63.15$} & $95.36$ & {$63.78$} \\
        \hline
        \hline
        Average & $63.14$  & {$35.29$} & $64.07$  & {$32.66$} & $62.93$  & {$32.66$} & $73.59$  & {$46.66$} & $72.06$  & {$46.03$} & $72.12$ & {$46.75$} \\
        
    \end{tabular}
    \caption{The performance of applying various \stepone to SSLSAM, weights acquired from \steptwo.}
    \label{tab:ap_sup}
\end{table*}

\begin{table*}[h]
    \centering
    \footnotesize
    \begin{tabular}{l|c|c|c|c|c|c|c|c|c|c|c|c}
        \multirow{4}{*}{Dataset} & \multicolumn{12}{c}{MedSAM} \\ \cline{2-13}
        & \multicolumn{6}{c|}{Decoder} & \multicolumn{6}{c}{En/Decoder} \\ \cline{2-13}
        & \multicolumn{2}{c|}{V}  & \multicolumn{2}{c|}{A} & \multicolumn{2}{c|}{L} & \multicolumn{2}{c|}{V} & \multicolumn{2}{c|}{A} & \multicolumn{2}{c}{L} \\ \cline{2-13}
        & DSC  & {NSD} & DSC  & {NSD} & DSC  & {NSD} & DSC  & {NSD} & DSC  & {NSD} & DSC  & {NSD}\\
        \hline
        MRI-Prost. & $71.40$  & {$18.05$} & $75.78$  & {$20.54$} & $70.82$  & {$15.44$} & $87.42$  & {$40.69$} & $89.76$  & {$48.84$} & $89.02$ & {$47.40$} \\
        MRI-Heart & $80.13$  & {$60.41$} & $81.90$  & {$64.03$} & $80.27$  & {$63.26$} & $84.97$  & {$73.29$} & $85.17$  & {$76.56$} & $85.59$ & {$75.33$} \\
        MRI-Brain & $61.76$  & {$25.07$} & $69.62$  & {$32.20$} & $60.33$  & {$23.47$} & $69.54$  & {$33.80$} & $80.60$  & {$54.07$} & $78.98$ & {$51.51$} \\
        MRI-Scler. & $22.20$  & {$18.29$} & $25.23$  & {$22.33$} & $25.16$  & {$22.49$} & $36.28$  & {$34.08$} & $43.49$  & {$43.14$} & $40.43$ & {$39.61$} \\
        MRI-Hippo. & $79.81$  & {$25.76$} & $81.69$  & {$28.10$} & $80.67$  & {$25.44$} & $82.70$  & {$30.19$} & $82.71$  & {$30.31$} & $82.67$ & {$29.24$} \\
        MRI-Kidney & $81.04$  & {$49.73$} & $82.28$  & {$49.14$} & $79.52$  & {$46.28$} & $87.95$  & {$61.80$} & $88.38$  & {$68.38$} & $70.84$ & {$69.14$} \\
        MRI-Stroke. & $60.21$  & {$46.98$} & $64.02$  & {$52.55$} & $62.38$  & {$48.93$} & $67.30$  & {$58.11$} & $72.09$  & {$60.98$} & $70.84$ & {$63.02$} \\
        MRI-Resect. & $51.53$  & {$26.41$} & $54.56$  & {$30.64$} & $49.31$  & {$24.98$} & $67.37$  & {$49.06$} & $76.41$  & {$63.31$} & $70.10$ & {$55.80$} \\
        MRI-Breast & $53.95$  & {$63.03$} & $54.80$  & {$62.91$} & $53.02$  & {$60.66$} & $61.93$  & {$70.06$} & $67.37$  & {$74.47$} & $66.64$ & {$74.32$} \\
        \hline
        CT-Colon & $21.84$  & {$13.58$} & $26.25$  & {$15.34$} & $22.54$  & {$14.78$} & $40.74$  & {$31.36$} & $55.23$  & {$44.76$} & $45.30$ & {$35.31$} \\
        CT-Panc. & $51.22$  & {$29.20$} & $59.16$  & {$35.69$} & $53.69$  & {$30.07$} & $66.07$  & {$47.49$} & $72.63$  & {$55.42$} & $72.26$ & {$52.42$} \\
        CT-Vessel & $40.19$  & {$43.35$} & $44.54$  & {$49.16$} & $39.74$  & {$41.14$} & $53.63$  & {$56.56$} & $57.83$  & {$59.06$} & $56.22$ & {$58.27$} \\
        \hline
        US-Breast & $57.55$  & {$14.65$} & $67.14$  & {$20.98$} & $62.34$  & {$18.36$} & $60.06$  & {$15.43$} & $76.88$  & {$33.15$} & $78.48$ & {$33.60$} \\
        US-Muscle & $84.22$  & {$19.43$} & $85.64$  & {$20.56$} & $84.11$  & {$18.53$} & $86.85$  & {$22.87$} & $86.90$  & {$24.33$} & $87.25$ & {$24.51$} \\
        US-Kidney & $96.05$  & {$58.25$} & $97.03$  & {$68.55$} & $97.18$  & {$61.07$} & $97.89$  & {$63.74$} & $98.18$  & {$81.37$} & $98.12$ & {$80.62$} \\
        \hline
        Xray-Hip & $95.52$  & {$62.12$} & $93.49$  & {$62.18$} & $93.27$  & {$59.53$} & $94.74$  & {$78.87$} & $94.73$  & {$77.58$} & $94.55$ & {$75.22$} \\
        Xray-Chest & $95.06$  & {$60.54$} & $95.28$  & {$61.65$} & $95.25$  & {$61.73$} & $95.79$  & {$69.20$} & $95.86$  & {$69.86$} & $95.77$ & {$69.92$} \\
        \hline
        \hline
        Average & $64.80$  & {$37.58$} & $68.14$  & {$41.22$} & $65.27$  & {$37.65$} & $73.14$  & {$49.52$} & $77.90$  & {$57.13$} & $75.47$ & {$55.33$} \\

    \end{tabular}
    \caption{The performance of applying various \stepone to MedSAM, weights acquired from \cite{MedSAM}.}
    \label{tab:ap_med}
\end{table*}

\begin{table*}[h]
    \centering
    \footnotesize
    \begin{tabular}{l|c|c|c|c|c|c|c|c|c|c|c|c}
        \multirow{4}{*}{Dataset} & \multicolumn{12}{c}{SSLSAM} \\ \cline{2-13}
        & \multicolumn{6}{c|}{Decoder} & \multicolumn{6}{c}{En/Decoder} \\ \cline{2-13}
        & \multicolumn{2}{c|}{V}  & \multicolumn{2}{c|}{A} & \multicolumn{2}{c|}{L} & \multicolumn{2}{c|}{V} & \multicolumn{2}{c|}{A} & \multicolumn{2}{c}{L} \\ \cline{2-13}
        & DSC  & {NSD} & DSC  & {NSD} & DSC  & {NSD} & DSC  & {NSD} & DSC  & {NSD} & DSC  & {NSD}\\
        \hline
        MRI-Prost. & $82.86$  & {$33.48$} & $84.38$  & {$34.08$} & $83.55$  & {$34.52$} & $88.82$  & {$44.53$} & $90.29$  & {$50.10$} & $86.74$ & {$39.23$} \\
        MRI-Heart & $78.25$  & {$58.08$} & $83.67$  & {$68.59$} & $84.08$  & {$69.83$} & $84.45$  & {$73.93$} & $87.41$  & {$78.75$} & $85.52$ & {$73.81$} \\
        MRI-Brain & $63.26$  & {$29.01$} & $71.22$  & {$36.88$} & $64.77$  & {$31.13$} & $76.71$  & {$50.13$} & $83.31$  & {$55.51$} & $80.33$ & {$51.39$} \\
        MRI-Scler. & $31.96$  & {$29.36$} & $32.32$  & {$29.82$} & $31.32$  & {$27.77$} & $39.80$  & {$40.00$} & $47.03$  & {$41.60$} & $37.74$ & {$35.98$} \\
        MRI-Hippo. & $80.63$  & {$30.00$} & $81.95$ & {$29.57$}  & $81.90$  & {$31.25$} & $83.13$  & {$40.52$} & $82.83$  & {$34.75$} & $82.28$ & {$32.71$} \\
        MRI-Kidney & $85.54$  & {$58.16$} & $87.10$  & {$61.05$} & $84.91$  & {$56.20$} & $89.80$  & {$64.29$} & $89.57$  & {$68.57$} & $89.69$ & {$67.00$} \\
        MRI-Stroke. & $65.10$  & {$52.10$} & $66.71$  & {$55.73$} & $63.80$  & {$48.91$} & $70.34$  & {$59.74$} & $73.89$  & {$62.33$} & $71.11$ & {$59.37$} \\
        MRI-Resect. & $67.92$  & {$49.55$} & $69.26$  & {$52.37$} & $69.98$  & {$51.64$} & $71.37$  & {$57.10$} & $78.73$  & {$66.47$} & $75.64$ & {$61.82$} \\
        MRI-Breast & $62.30$  & {$70.34$} & $62.15$  & {$70.77$} & $60.97$  & {$69.03$} & $66.23$  & {$73.68$} & $69.72$  & {$75.92$} & $67.65$ & {$73.66$} \\
        \hline
        CT-Colon & $35.31$  & {$23.78$} & $37.70$  & {$8.23$} & $23.44$  & {$10.42$} & $45.29$  & {$41.50$} & $55.56$ & {$45.50$}  & $45.49$ & {$35.51$} \\
        CT-Panc. & $58.99$  & {$35.69$} & $66.19$  & {$42.16$} & $61.29$ & {$38.58$}  & $72.07$ & {$53.84$}  & $74.50$ & {$57.05$}  & $70.66$ & {$50.63$} \\
        CT-Vessel & $48.49$ & {$49.78$}  & $47.76$  & {$51.46$} & $49.97$  & {$49.68$} & $42.14$  & {$53.52$} & $57.67$  & {$59.99$} & $57.51$ & {$57.86$} \\
        \hline
        US-Breast & $56.43$  & {$15.21$} & $68.46$  & {$22.30$} & $63.22$  & {$18.58$} & $78.15$  & {$33.97$} & $77.84$  & {$33.31$} & $75.16$ & {$28.54$} \\
        US-Muscle & $84.26$  & {$19.79$} & $85.76$  & {$19.31$} & $85.33$  & {$20.29$} & $85.67$  & {$24.60$} & $87.70$  & {$22.64$} & $86.99$ & {$23.00$} \\
        US-Kidney & $96.64$  & {$64.48$} & $97.34$  & {$70.84$} & $96.78$  & {$65.26$} & $98.02$ & {$81.94$}  & $98.24$  & {$82.39$} & $97.96$ & {$78.80$} \\
        \hline
        Xray-Hip & $93.73$  & {$65.77$} & $93.18$  & {$59.00$} & $93.59$  & {$63.39$} & $94.14$  & {$23.03$} & $94.73$  & {$76.89$} & $93.96$ & {$67.01$} \\
        Xray-Chest & $95.32$  & {$63.68$} & $95.04$  & {$58.85$} & $95.23$  & {$62.57$} & $94.93$  & {$52.72$} & $95.72$  & {$66.90$} & $95.80$ & {$68.38$} \\
        \hline
        \hline
        Average & $69.82$  & {$44.28$} & $72.36$  & {$45.61$} & $70.24$  & {$44.32$} & $75.36$  & {$51.32$} & $79.10$  & {$57.95$} & $76.48$ & {$53.51$} \\

    \end{tabular}
    \caption{The performance of applying various \stepone to SSLSAM, weights acquired from \stepthree.}
    \label{tab:ap_ssl}
\end{table*}

\begin{table*}[h]
    \centering
    \footnotesize
    \begin{tabular}{l|c|c|c|c|c|c|c|c|c|c|c|c}
        \multirow{4}{*}{Dataset} & \multicolumn{4}{c|}{ViT-B} & \multicolumn{4}{c|}{MedSAM} & \multicolumn{4}{c}{SSLSAM}\\ \cline{2-13}
        & \multicolumn{2}{c|}{Decoder} & \multicolumn{2}{c|}{En/Decoder} & \multicolumn{2}{c|}{Decoder} & \multicolumn{2}{c|}{En/Decoder} & \multicolumn{2}{c|}{Decoder} & \multicolumn{2}{c}{En/Decoder}\\ \cline{2-13}
        & V & A & V & A & V & A & V & A & V & A & V & A \\
        \hline
        & \multicolumn{12}{c}{DSC} \\ \cline{2-13}
        MRI-Prost. & $66.77$  & $68.15$  & $45.47$  & $73.11$  & $58.73$ & $57.93$  & $73.04$  & $70.20$  & $62.69$& $67.46$   & $69.67$ & $73.88$\\
        MRI-Heart & $56.94$  & $62.55$  & $19.14$  & $72.34$  & $64.70$  & $59.65$  & $69.21$  & $66.28$  & $69.00$  & $68.90$  &$35.30$ & $61.30$  \\
        MRI-Brain & $31.54$  & $38.29$  & $3.28$  & $60.31$  & $17.10$  & $43.01$  & $41.33$  & $64.50$  &  $31.62$ & $50.19$  & $0.09$  & $55.09$  \\
        MRI-Scler. & $2.34$  & $1.49$  & $2.26$ & $16.17$   & $4.17$   & $3.58$  & $13.08$  & $12.91$  & $2.59$ & $2.39$    & $0.98$ & $12.70$   \\
        MRI-Hippo. & $60.12$  & $51.86$  & $70.45$  & $67.79$  & $46.23$ & $51.93$  & $65.05$  & $60.49$  & $42.91$  & $51.42$  & $0.00$ & $64.73$  \\
        MRI-Kidney & $74.46$  & $74.45$  & $67.70$  & $76.21$  & $68.40$ & $66.98$  & $79.27$  & $67.56$  & $73.90$  & $66.70$  & $0.00$ & $77.10$  \\
        MRI-Stroke. & $8.59$  & $15.46$  & $25.20$  & $25.54$  & $11.60$ & $15.79$  & $27.09$  & $34.02$  & $14.03$ & $26.60$  & $31.60$ & $25.50$  \\
        MRI-Resect. & $27.01$ & $35.33$  & $11.30$ & $51.30$   & $13.90$ & $17.75$  & $29.38$  & $33.00$  & $36.81$  & $46.40$  & $4.44$ & $50.30$  \\
        MRI-Breast & $45.00$  & $43.83$  & $33.61$  & $52.25$  & $36.30$ & $37.60$  & $38.12$  & $42.27$  & $40.17$ & $42.50$  & $21.00$ & $48.80$  \\
        \hline
        \hline
        Average & $41.42$  & $43.49$  & $36.05$  & $55.01$  & $35.83$  & $39.36$  & $48.37$  & $50.35$  & $41.52$  & $46.95$  & $18.12$ & $52.16$  \\
        \hline
        & \multicolumn{12}{c}{{NSD}} \\ \cline{2-13}
        MRI-Prost. & {$19.37$}  & {$18.60$}  & {$1.21$}  & {$23.61$}  & {$7.96$}  & {$9.52$}  & {$21.55$}  & {$19.49$}  & {$13.99$}  & {$14.48$}  & {$6.08$}  & {$20.06$} \\
        MRI-Heart & {$31.09$}  & {$41.58$}  & {$0.08$}  & {$50.52$}  & {$41.53$}  & {$27.22$}  & {$45.22$}  & {$42.02$}  & {$42.70$}  & {$40.51$}  & {$0.00$}  & {$34.04$} \\
        MRI-Brain & {$7.76$}  & {$11.94$}  & {$3.95$}  & {$21.71$}  & {$5.30$}  & {$13.17$}  & {$10.57$}  & {$32.38$}  & {$8.00$}  & {$15.63$}  & {$1.27$}  & {$19.34$} \\
        MRI-Scler. & {$1.75$}  & {$1.62$}  & {$3.33$}  & {$17.00$}  & {$2.92$}  & {$2.87$}  & {$11.14$}  & {$14.34$}  & {$1.16$}  & {$1.89$}  & {$7.56$}  & {$8.76$} \\
        MRI-Hippo. & {$12.03$}  & {$10.64$}  & {$7.80$}  & {$23.79$}  & {$8.31$}  & {$10.24$}  & {$13.90$}  & {$12.49$}  & {$5.77$}  & {$10.84$}  & {$0.76$}  & {$13.63$} \\
        MRI-Kidney & {$44.17$}  & {$42.86$}  & {$0.00$}  & {$51.76$}  & {$23.77$}  & {$25.00$}  & {$43.70$}  & {$28.31$}  & {$39.61$}  & {$32.26$}  & {$0.00$}  & {$45.29$} \\
        MRI-Stroke. & {$7.25$}  & {$10.52$}  & {$18.93$}  & {$27.77$}  & {$8.64$}  & {$10.39$}  & {$18.03$}  & {$26.07$}  & {$9.92$}  & {$18.27$}  & {$0.00$}  & {$16.55$} \\
        MRI-Resect. & {$12.64$}  & {$14.36$}  & {$2.52$}  & {$27.63$}  & {$6.18$}  & {$8.89$}  & {$7.08$}  & {$11.89$}  & {$13.95$}  & {$19.36$}  & {$1.01$}  & {$21.38$} \\
        MRI-Breast & {$50.72$}  & {$49.82$}  & {$31.22$}  & {$58.87$}  & {$38.63$}  & {$41.03$}  & {$39.66$}  & {$46.58$}  & {$45.33$}  & {$48.69$}  & {$1.16$}  & {$52.98$} \\
        \hline
        \hline
        Average & {$20.83$}  & {$22.52$}  & {$7.68$}  & {$33.63$}  & {$15.97$}  & {$16.53$}  & {$23.52$}  & {$26.03$}  & {$20.12$}  & {$22.51$}  & {$1.99$}  & {$25.87$} \\
    \end{tabular}
    \caption{The performance of applying \stepone to the ViT-B architecture on the few-shot setting {5-shot} with different initial weights.}
    \label{tab:ap_fs_sam}
\end{table*}

\begin{table*}[h]
    \centering
    \begin{tabular}{l|c|c|c|c|c|c}
        \multirow{2}{*}{Dataset} & \multicolumn{2}{c|}{UNet} & \multicolumn{2}{c|}{nnUNet} & \multicolumn{2}{c}{SwinUNet}\\ \cline{2-7}
        & DSC  & {NSD} & DSC  & {NSD} & DSC  & {NSD} \\
        \hline
        MRI-Prost. & $6.34$  & {$1.43$}  & $70.13$  & {$22.75$}  & $2.67$ & {$1.26$}  \\
        MRI-Heart & $1.26$  & {$4.14$}  & $67.69$  & {$55.95$}  & $59.94$ & {$34.23$}  \\
        MRI-Brain & $6.49$  & {$1.32$}  & $46.60$  & {$25.78$}  & $53.69$ & {$18.20$}  \\
        MRI-Scler. & $0.77$  & {$1.25$}  & $10.40$  & {$9.35$}  & $1.64$ & {$0.37$}  \\
        MRI-Hippo. & $34.05$  & {$4.39$}  & $71.29$  & {$22.34$}  & $46.58$ & {$7.76$}  \\
        MRI-Kidney & $47.95$  & {$21.63$}  & $75.81$  & {$48.88$}  & $47.51$ & {$18.00$}  \\
        MRI-Stroke. & $2.01$  & {$4.67$}  & $37.30$  & {$26.20$}  & $38.99$ & {$23.36$}  \\
        MRI-Resect. & $2.66$  & {$2.50$}  & $54.34$  & {$38.29$}  & $22.16$ & {$4.63$}  \\
        MRI-Breast & $32.40$ & {$35.14$}  & $54.06$  & {$58.97$}  & $37.39$ & {$37.79$}  \\
        \hline
        \hline
        Average & $14.88$  & {$8.50$}  & $54.18$  & {$34.28$}  & $34.51$ & {$16.18$}  \\

    \end{tabular}    
    \caption{The performance of UNet, nnUNet, and SwinUNet across different datasets under a few-shot setting {(5-shot)}.}
    \label{tab:ap_fs_unet}
\end{table*}

\begin{table}[h]
    \centering
    \footnotesize
    \begin{tabular}{l|c|c|c|c|c|c}
        \multirow{2}{*}{Dataset} & \multicolumn{2}{c|}{ViT-H} & \multicolumn{2}{c|}{ViT-B} & \multicolumn{2}{c}{ViT-T}\\ \cline{2-7}
        & H & O & H & O& H & O \\
        \hline
        MRI-Prost. & $52.63$  & $67.65$  & $35.52$  & $68.94$  & $41.06$  & $65.71$ \\
        MRI-Heart & $49.20$  & $58.75$  & $48.39$  & $65.24$  & $43.50$  & $63.42$ \\
        MRI-Brain & $34.20$  & $54.57$  & $29.23$  & $58.55$  & $30.33$  & $58.38$ \\
        MRI-Scler. & $1.80$  & $3.91$  & $1.58$  & $5.28$  & $1.72$  & $4.50$ \\
        MRI-Hippo. & $34.78$  & $55.15$  & $17.34$  & $59.33$  & $22.20$  & $59.94$ \\
        MRI-Kidney & $41.41$  & $73.31$  & $44.17$  & $76.22$  & $46.07$  & $70.85$ \\
        MRI-Stroke. & $21.38$  & $39.85$  & $21.75$  & $49.98$  & $20.17$  & $44.43$ \\
        MRI-Resect. & $27.86$  & $45.84$  & $17.38$  & $60.82$  & $26.07$  & $55.89$ \\
        MRI-Breast & $5.00$  & $21.61$  & $12.83$  & $21.81$  & $5.00$  & $28.57$ \\
        \hline
        \hline
        Average & $29.81$  & $46.74$  & $25.35$  & $51.80$  & $26.24$  & $50.19$ \\
    \end{tabular}
    \caption{The performance of SAM with different choices of encoder architectures with point prompts. H = Highest confidence, O = Oracle mode.}
    \label{tab:ap_point_ori}
\end{table}

\begin{table}[h]
    \centering
    \footnotesize
    \begin{tabular}{l|c|c|c|c|c|c}
        \multirow{3}{*}{Dataset} & \multicolumn{6}{c}{ViT-H} \\ \cline{2-7}
        & \multicolumn{3}{c|}{Decoder} & \multicolumn{3}{c}{En/Decoder} \\\cline{2-7}
        & V & A & L & V & A & L \\
        \hline
        MRI-Prost. & $87.74$  & $86.89$  & $86.82$  & $79.84$  & $89.55$  & $88.10$ \\
        MRI-Heart & $84.77$  & $85.07$  & $86.46$  & $83.77$  & $85.60$  & $86.40$ \\
        MRI-Brain & $74.44$  & $72.87$  & $72.66$  & $74.15$  & $80.76$  & $75.09$ \\
        MRI-Scler. & $62.95$  & $62.64$  & $62.64$  & $51.42$  & $60.77$  & $64.26$ \\
        MRI-Hippo. & $83.13$  & $82.03$  & $82.95$  & $83.99$  & $81.88$  & $83.26$ \\
        MRI-Kidney & $88.49$  & $87.37$  & $89.06$  & $89.31$  & $86.87$  & $92.01$ \\
        MRI-Stroke. & $61.18$  & $66.09$  & $60.47$  & $73.87$  & $78.22$  & $81.52$ \\
        MRI-Resect. & $73.38$  & $75.20$  & $68.06$  & $66.57$  & $79.96$  & $77.09$ \\
        MRI-Breast & $64.84$  & $69.56$  & $67.61$  & $70.12$  & $77.14$  & $78.13$ \\
        \hline
        \hline
        Average & $75.66$  & $76.41$  & $75.19$  & $74.78$  & $80.08$  & $80.65$ \\

    \end{tabular}
    \caption{The performance of applying \stepone during point-based segmentation with ViT-H as the encoder architecture.}
    \label{tab:ap_pointh}
\end{table}

\begin{table}[h]
    \centering
    \footnotesize
    \begin{tabular}{l|c|c|c|c|c|c}
        \multirow{3}{*}{Dataset} & \multicolumn{6}{c}{ViT-B} \\ \cline{2-7}
        & \multicolumn{3}{c|}{Decoder} & \multicolumn{3}{c}{En/Decoder} \\\cline{2-7}
        & V & A & L & V & A & L \\
        \hline
        MRI-Prost. & $87.23$  & $87.22$  & $87.10$  & $88.54$  & $89.57$  & $87.48$ \\
        MRI-Heart & $84.90$  & $85.20$  & $84.98$  & $86.04$  & $87.54$  & $85.41$ \\
        MRI-Brain & $76.05$  & $76.48$  & $77.71$  & $73.82$  & $83.70$  & $77.61$ \\
        MRI-Scler. & $58.41$  & $59.18$  & $62.78$  & $59.42$  & $66.95$  & $60.98$ \\
        MRI-Hippo. & $82.06$  & $82.98$  & $83.66$  & $84.23$  & $84.04$  & $83.43$ \\
        MRI-Kidney & $89.34$  & $89.04$  & $86.93$  & $88.81$  & $90.80$  & $89.77$ \\
        MRI-Stroke. & $69.08$  & $69.67$  & $52.63$  & $72.62$  & $79.19$  & $78.12$ \\
        MRI-Resect. & $73.11$  & $77.15$  & $73.02$  & $71.36$  & $81.56$  & $75.39$ \\
        MRI-Breast & $64.57$  & $67.79$  & $69.79$  & $56.86$  & $73.38$  & $76.28$ \\
        \hline
        \hline
        Average & $76.08$  & $77.19$  & $75.40$  & $75.74$  & $81.86$  & $79.39$ \\       
    \end{tabular}
    \caption{The performance of applying \stepone during point-based segmentation with ViT-B as the encoder architecture.}
    \label{tab:ap_pointb}
\end{table}

\begin{table}[h]
    \centering
    \footnotesize
    \begin{tabular}{l|c|c|c|c|c|c}
        \multirow{3}{*}{Dataset} & \multicolumn{6}{c}{ViT-T} \\ \cline{2-7}
        & \multicolumn{3}{c|}{Decoder} & \multicolumn{3}{c}{En/Decoder} \\\cline{2-7}
        & V & A & L & V & A & L \\
        \hline
        MRI-Prost. & $87.76$  & $84.06$  & $78.06$  & $90.46$  & $90.81$  & $84.38$ \\
        MRI-Heart & $85.19$  & $85.50$  & $79.28$  & $87.72$  & $88.03$  & $86.60$ \\
        MRI-Brain & $74.38$  & $72.63$  & $59.00$  & $84.12$  & $85.78$  & $83.98$ \\
        MRI-Scler. & $61.29$  & $63.85$  & $50.53$  & $64.78$  & $72.56$  & $71.04$ \\
        MRI-Hippo. & $82.97$  & $80.71$  & $79.20$  & $84.95$  & $87.36$  & $77.90$ \\
        MRI-Kidney & $88.30$  & $87.60$  & $81.25$  & $91.12$  & $91.48$  & $89.69$ \\
        MRI-Stroke. & $63.80$  & $65.78$  & $64.84$  & $76.22$  & $81.61$  & $81.21$ \\
        MRI-Resect. & $74.08$  & $74.48$  & $68.94$  & $80.66$  & $81.62$  & $79.82$ \\
        MRI-Breast & $59.92$  & $64.80$  & $59.12$  & $69.55$  & $75.23$  & $73.61$ \\
        \hline
        \hline
        Average & $75.30$  & $75.49$  & $68.91$  & $81.06$  & $83.83$  & $80.91$ \\

    \end{tabular}
    \caption{The performance of applying \stepone during point-based segmentation with ViT-T as the encoder architecture.}
    \label{tab:ap_pointt}
\end{table}

\begin{table*}[h]
    \centering
    \footnotesize
    \begin{tabular}{l|c|c|c|c|c|c|c|c|c}
        \multirow{2}{*}{Dataset} & \multicolumn{2}{c|}{ViT-H} & \multicolumn{2}{c|}{ViT-B} & \multicolumn{2}{c|}{ViT-T} & \multicolumn{2}{c|}{MedSAM} & SupSAM \\ \cline{2-10}
        & H & O & H & O& H & O & H & O & - \\
        \hline
        MRI-Prost. & $87.09$  & $91.60$  & $89.75$  & $92.20$  & $88.50$  & $90.99$  & $87.02$  & $92.51$  & $91.82$ \\
        MRI-Heart & $67.29$  & $85.62$  & $83.03$  & $85.96$  & $83.49$  & $86.82$  & $73.99$  & $75.73$  & $83.74$ \\
        MRI-Brain & $73.44$  & $84.51$  & $81.74$  & $85.48$  & $81.70$  & $84.50$  & $77.13$  & $78.32$  & $80.57$ \\
        MRI-Scler. & $39.13$  & $78.58$  & $75.15$  & $79.91$  & $76.43$  & $79.72$  & $38.52$  & $41.28$  & $56.26$ \\
        MRI-Hippo. & $71.19$  & $81.37$  & $78.48$  & $82.31$  & $78.52$  & $81.75$  & $70.76$  & $72.69$  & $84.75$ \\
        MRI-Kidney & $87.51$  & $89.58$  & $87.84$  & $90.02$  & $87.40$  & $89.31$  & $70.36$  & $72.44$  & $87.46$ \\
        MRI-Stroke. & $55.97$  & $81.09$  & $78.14$  & $82.21$  & $77.50$  & $82.15$  & $55.88$  & $57.95$  & $78.06$ \\
        MRI-Resect. & $71.97$  & $85.68$  & $83.58$  & $87.09$  & $81.58$  & $84.85$  & $66.92$  & $68.20$  & $78.90$ \\
        MRI-Breast & $54.69$  & $63.38$  & $56.87$  & $66.96$  & $54.09$  & $63.06$  & $19.52$  & $21.12$  & $64.70$ \\
        \hline
        \hline
        Average & $67.59$  & $82.38$  & $79.40$  & $83.57$  & $78.80$  & $82.57$  & $62.23$  & $64.47$  & $78.47$ \\

    \end{tabular}
    \caption{The performance of SAM with different choices of encoder architectures with box prompts. H = Highest confidence, O = Oracle mode. MedSAM and SupSAM is applicable since both utilize box during training.}
    \label{tab:ap_box_ori}
\end{table*}

\begin{table}[h]
    \centering
    \footnotesize
    \begin{tabular}{l|c|c|c|c|c|c}
        \multirow{3}{*}{Dataset} & \multicolumn{6}{c}{ViT-H} \\ \cline{2-7}
        & \multicolumn{3}{c|}{Decoder} & \multicolumn{3}{c}{En/Decoder} \\\cline{2-7}
        & V & A & L & V & A & L \\
        \hline
        MRI-Prost. & $91.61$  & $94.18$  & $94.52$  & $91.08$  & $94.82$  & $95.16$ \\
        MRI-Heart & $64.24$  & $90.66$  & $88.91$  & $86.21$  & $89.17$  & $91.88$ \\
        MRI-Brain & $77.84$  & $82.93$  & $83.21$  & $73.03$  & $87.09$  & $88.14$ \\
        MRI-Scler. & $62.29$  & $73.70$  & $74.06$  & $62.53$  & $74.84$  & $72.18$ \\
        MRI-Hippo. & $84.95$  & $86.02$  & $88.32$  & $88.14$  & $88.61$  & $89.45$ \\
        MRI-Kidney & $89.44$  & $89.59$  & $90.06$  & $89.51$  & $90.06$  & $90.15$ \\
        MRI-Stroke. & $81.05$  & $81.14$  & $81.95$  & $81.66$  & $87.70$  & $86.42$ \\
        MRI-Resect. & $81.66$  & $81.29$  & $81.47$  & $83.23$  & $86.93$  & $87.70$ \\
        MRI-Breast & $72.91$  & $74.71$  & $72.89$  & $69.55$  & $79.70$  & $80.68$ \\
        \hline
        \hline
        Average & $78.44$  & $83.80$  & $83.93$  & $80.55$  & $86.55$  & $86.86$ \\

    \end{tabular}
    \caption{The performance of applying \stepone during box-based segmentation with ViT-H as the encoder architecture.}
    \label{tab:ap_boxh}
\end{table}

\begin{table}[h]
    \centering
    \footnotesize
    \begin{tabular}{l|c|c|c|c|c|c}
        \multirow{3}{*}{Dataset} & \multicolumn{6}{c}{ViT-B} \\ \cline{2-7}
        & \multicolumn{3}{c|}{Decoder} & \multicolumn{3}{c}{En/Decoder} \\\cline{2-7}
        & V & A & L & V & A & L \\
        \hline
        MRI-Prost. & $94.13$  & $94.43$  & $94.76$  & $92.76$  & $94.78$  & $94.58$ \\
        MRI-Heart & $90.27$  & $90.52$  & $90.35$  & $90.18$  & $90.48$  & $90.05$ \\
        MRI-Brain & $82.68$  & $82.85$  & $84.71$  & $76.32$  & $87.98$  & $85.27$ \\
        MRI-Scler. & $68.58$  & $71.80$  & $69.12$  & $61.74$  & $69.43$  & $72.00$ \\
        MRI-Hippo. & $86.38$  & $87.02$  & $89.04$  & $85.89$  & $88.24$  & $88.68$ \\
        MRI-Kidney & $91.37$  & $90.33$  & $89.44$  & $91.63$  & $91.53$  & $90.89$ \\
        MRI-Stroke. & $72.45$  & $81.19$  & $75.46$  & $67.33$  & $85.84$  & $82.76$ \\
        MRI-Resect. & $77.15$  & $77.35$  & $83.83$  & $76.86$  & $84.91$  & $85.49$ \\
        MRI-Breast & $64.57$  & $69.88$  & $73.70$  & $59.66$  & $74.00$  & $73.14$ \\
        \hline
        \hline
        Average & $80.84$  & $82.82$  & $83.38$  & $78.04$  & $85.24$  & $84.76$ \\

    \end{tabular}
    \caption{The performance of applying \stepone during box-based segmentation with ViT-B as the encoder architecture.}
    \label{tab:ap_boxb}
\end{table}

\begin{table}[h]
    \centering
    \footnotesize
    \begin{tabular}{l|c|c|c|c|c|c}
        \multirow{3}{*}{Dataset} & \multicolumn{6}{c}{ViT-T} \\ \cline{2-7}
        & \multicolumn{3}{c|}{Decoder} & \multicolumn{3}{c}{En/Decoder} \\\cline{2-7}
        & V & A & L & V & A & L \\
        \hline
        MRI-Prost. & $93.37$  & $94.45$  & $94.07$  & $93.51$  & $94.98$  & $94.13$ \\
        MRI-Heart & $88.75$  & $91.26$  & $90.10$  & $90.50$  & $91.02$  & $91.05$ \\
        MRI-Brain & $78.92$  & $82.23$  & $79.38$  & $84.15$  & $88.69$  & $88.01$ \\
        MRI-Scler. & $75.41$  & $76.24$  & $74.59$  & $74.29$  & $79.49$  & $78.49$ \\
        MRI-Hippo. & $86.64$  & $88.01$  & $87.59$  & $87.60$  & $90.38$  & $86.85$ \\
        MRI-Kidney & $90.27$  & $89.37$  & $87.13$  & $92.15$  & $92.33$  & $91.22$ \\
        MRI-Stroke. & $70.94$  & $81.77$  & $81.40$  & $81.44$  & $89.33$  & $88.13$ \\
        MRI-Resect. & $76.44$  & $85.68$  & $82.91$  & $81.00$  & $86.77$  & $87.62$ \\
        MRI-Breast & $62.81$  & $70.90$  & $63.12$  & $74.09$  & $80.54$  & $74.74$ \\
        \hline
        \hline
        
        Average & $80.39$  & $84.43$  & $82.25$  & $84.30$  & $88.17$  & $86.69$ \\

    \end{tabular}
    \caption{The performance of applying \stepone during box-based segmentation with ViT-T as the encoder architecture.}
    \label{tab:ap_boxt}
\end{table}

\end{document}